\documentclass[lettersize,journal]{IEEEtran}
\ifCLASSINFOpdf
\else
\fi
\usepackage{cite}
\usepackage{amsmath,amssymb,amsfonts}
\usepackage{algorithm}
\usepackage{algpseudocode}
\usepackage{graphicx}
\usepackage{textcomp}
\usepackage{xcolor}
\usepackage{float}
\usepackage{booktabs} 
\usepackage{hyperref}
\usepackage{url}

\algrenewcommand\algorithmiccomment[1]{\hfill\(\triangleright\) #1}
\algrenewcommand\alglinenumber[1]{\scriptsize #1} 

\begin{document}
%
\title{Adaptive Multi-Discriminator WGAN Framework for Resource-Constrained Internet of Vehicles Using Reinforcement Learning and Game Theory}
%
%
%

\author{Farhoud~Jafari Kaleibar, Amr M. Zaki~\IEEEmembership{Member,~IEEE,} and        Marin~Litoiu,~\IEEEmembership{Senior Member,~IEEE,}
\thanks{F. Jafari Kaleibar, Amr M. Zaki and M. Litoiu are with the Department
of Electrical and Computer Engineering, York University, Toronto,
ON, Canada, e-mail: (farhoud, mlitoiu@yorku.ca)}
}

\maketitle

\begin{abstract}
Managing machine learning workloads as a network service introduces a resource-orchestration problem distinct from conventional model training; which nodes should be allocated to a task, how communication and computation budgets should be divided among them, and how service quality should be sustained as connectivity and node availability change with mobility. Deploying Generative Adversarial Networks (GANs) in Internet of Vehicles (IoV) environments is a demanding instance of this problem; resource constraints, dynamic network topologies, and competing optimization objectives mean that traditional GAN architectures cannot simultaneously achieve high accuracy, efficient resource use, low delay, and low communication overhead. This paper introduces an
adaptive multi-discriminator Wasserstein GAN (MD-WGAN) framework
that integrates reinforcement learning with game-theoretic
coordination to address these challenges jointly. In our framework,
roadside units host generators paired with Deep Q-Network (DQN)
agents that select discriminator subsets and manage distributed
training across mobile vehicular nodes, while a game-theoretic
coordination step allocates training epochs between generators and
discriminators. A unified optimization objective ties adversarial
learning quality to resource efficiency, communication overhead, and
latency under vehicular constraints, allowing the framework to
continuously adapt its training behavior as network conditions
change. Evaluation on real-world NGSIM trajectory data shows that the framework attains prediction accuracy comparable to state-of-the-art GAN baselines — the lowest RMSE (1.029) and MAE (0.894) among all evaluated methods — while markedly improving resource efficiency: average CPU utilization is reduced by roughly 28\% and mean memory usage by roughly 6\%, at competitive communication overhead and latency.
\end{abstract}

\begin{IEEEkeywords}
Network resource management, service orchestration, adaptive resource allocation, reinforcement learning, multi-agent coordination, generative adversarial networks, Internet of Vehicles, mobile edge computing.
\end{IEEEkeywords}

%
\IEEEpeerreviewmaketitle

\section{Introduction}
%
%
%
%
\IEEEPARstart{M}{anaging} and orchestrating machine learning services across mobile and edge network infrastructure is an emerging network management challenge, requiring adaptive allocation of computational and communication resources as devices collaborate under mobility and resource constraints~\cite{pratap2023cotev}. Generative
Adversarial Networks (GANs) have emerged as powerful tools within
this ecosystem, supporting traffic pattern synthesis, sensor data
augmentation, and predictive modeling for autonomous driving
systems~\cite{devika2024vadgan,zhang2023novel}. A central obstacle in
developing such IoV applications is the scarcity, cost, and privacy
sensitivity of real-world vehicular data~\cite{mishra2023collaborative,farhoudtsc}:
collecting sufficient labeled data for rare or safety-critical
scenarios (e.g., near-collision events, adverse weather) is often
impractical or dangerous, and sharing raw sensor and trajectory data
raises privacy concerns tied to individual location and driving
behavior. By learning the underlying distribution of real vehicular
data, GANs can synthesize additional training samples that augment
scarce datasets, reproduce rare or hazardous scenarios for safe
offline testing, and support privacy-preserving data sharing, since
synthetic samples can be exchanged in place of raw sensor
data~\cite{mfogo2025gan,luo2025high,amrvec}. 

However, deploying GANs in vehicular networks introduces challenges
distinct from conventional settings, stemming from their highly
mobile, resource-constrained, and latency-sensitive
nature~\cite{contreras2025generative,jafari2021sla,hammoud2022dynamic}.
Traditional single-generator, single-discriminator architectures
struggle under these conditions: intermittent connectivity and
limited edge-device resources exacerbate training instability, mode
collapse, and convergence difficulties, particularly as data
distributions shift with vehicle mobility and network
availability~\cite{yuan2020dynamic,karim2022architecture,elbir2022federated}.
The distributed nature of IoV (spanning V2V, V2I, and V2C
communication) further complicates coordination across
heterogeneous nodes with varying computational capability and
bandwidth~\cite{wang2022joint,posner2021federated,kaleibar2025customized}.
Existing approaches typically optimize individual metrics in
isolation, rather than jointly addressing latency, accuracy,
resource utilization, and network overhead as competing
objectives~\cite{wang2022federated,wu2024federated,xie2024gai}.

Multi-agent architectures offer a promising direction, exploiting
the distributed structure of vehicular networks to enable
collaborative learning under mobility and intermittent
connectivity~\cite{quan2024federated,hazarika2023multi,xu2024multi,Farhoud2025MDGAN}.
Coordinating such a system, however, is itself a multi-objective
problem: training efficiency, accuracy, resource allocation, and
communication overhead are all in tension under vehicle mobility,
heterogeneous device capabilities, and handover constraints.
Reinforcement learning, particularly Q-learning, offers a natural
coordination mechanism, enabling adaptive policy learning for
resource allocation and training as network conditions
change~\cite{liang2020reinforcement,zaki2025cogroup}; combined with
the improved training stability of Wasserstein GAN
(WGAN)~\cite{WGAN}, this allows the system to jointly optimize
training and resource metrics as mobility and computational load
evolve~\cite{hua2019gan}.

To address these challenges, this paper introduces an adaptive
multi-discriminator, multi-generator WGAN framework (MD-WGAN) for
IoV applications, using Q-learning to jointly optimize latency,
accuracy, resource efficiency, and network overhead under vehicular
mobility and edge computing constraints. We map the generator and
discriminator roles onto the physical IoV topology -- stationary,
resource-rich roadside units (RSUs) act as generators, while mobile
vehicles act as discriminators evaluating generated data under
mobility constraints (Section~\ref{systemmodel}) -- and formulate
discriminator selection and resource coordination as a multi-agent
reinforcement learning problem, complemented by a game-theoretic
coordination step that allocates training epochs between generators
and discriminators to improve convergence under network dynamics.

The primary contributions of this work are:
\begin{itemize}
    \item An adaptive multi-discriminator Wasserstein GAN framework
    for mobile and edge computing that integrates reinforcement
    learning-based discriminator selection with game-theoretic
    training coordination to address the challenges of dynamic
    vehicular networks;

    \item A unified problem formulation consolidating adversarial
    learning, adaptive resource management, communication
    efficiency, and multi-level convergence detection into a single
    constrained optimization framework;

    \item A hierarchical coordination protocol combining Deep
    Q-Network (DQN) agents for discriminator subset selection with
    game-theoretic best-response coordination for training epoch
    allocation, enabling the system to adapt to network conditions,
    resource availability, and data distribution shifts; and

    \item A comprehensive empirical evaluation demonstrating improved
    convergence efficiency, resource utilization, and model quality
    compared to other approaches, with analysis of
    scalability across varying vehicle densities.
\end{itemize}

The rest of the paper is organized as follows: Section~\ref{related}
provides background on GANs, WGAN, and IoV constraints, and reviews
related work on generative AI in distributed and vehicular settings.
Section~\ref{systemmodel} presents the system architecture and
network model, while Section~\ref{framework} presents the adaptive
optimization framework and problem formulation. Section~\ref{simulation}
discusses experiments and performance evaluation. Finally,
Section~\ref{conclusion} concludes the paper and outlines future
research directions.

\section{Background and Related Work} \label{related}

\subsection{Background}

\subsubsection{Generative Adversarial Networks}
A GAN~\cite{gan2018} consists of two competing neural networks trained
adversarially: a \emph{generator}, which learns to synthesize data
samples (e.g., synthetic vehicular traffic patterns) from a latent
random input, and a \emph{discriminator}, which learns to
distinguish these synthesized samples from real data. The two
networks are trained jointly: the generator improves by learning
to fool the discriminator, while the discriminator improves by
learning to detect increasingly realistic synthetic samples,
driving both toward higher-quality data generation over successive
training rounds.

\subsubsection{Wasserstein GAN}
Traditional GANs are trained by minimizing an approximation of the
Jensen--Shannon divergence between the real and generated data
distributions~\cite{gan2018}. This formulation is prone to training instability,
vanishing gradients when the discriminator becomes too confident,
and mode collapse, where the generator produces a limited variety
of outputs rather than capturing the full diversity of the real
data distribution. The Wasserstein GAN (WGAN)~\cite{WGAN}
addresses these issues by instead minimizing the Wasserstein
(Earth Mover's) distance between the two distributions, which
measures the minimum cost of transforming one distribution into
the other. Unlike the Jensen--Shannon divergence, the Wasserstein
distance provides a smooth, meaningful gradient signal even when
the real and generated distributions have little or no overlap --
a common situation early in training or with highly heterogeneous
data such as vehicular traffic patterns -- resulting in more stable
convergence. In place of a discriminator that outputs a bounded
real/fake probability, WGAN uses a critic that outputs an unbounded
real-valued score, constrained to be 1-Lipschitz continuous; this
constraint is enforced in practice via a gradient penalty term, as
reflected in the loss formulation of Eq.~(1). For consistency with
the IoV systems terminology used throughout this paper, we continue
to refer to these WGAN critics as ``discriminators.''

\subsubsection{Internet of Vehicles Characteristics}
The IoV environment introduces constraints not present in
conventional GAN training settings: vehicles connect and disconnect
intermittently as they move through RSU coverage zones, available
computational and communication resources vary across heterogeneous
devices, and training must proceed despite these fluctuating
conditions without centralized, always-on infrastructure~\cite{malektaji2021deep}. These
constraints motivate the distributed, resource-aware, multi-discriminator
design introduced in Section~\ref{systemmodel}.

\subsection{Related Work}
The rapid evolution of the IoV has necessitated the development of
sophisticated artificial intelligence frameworks capable of
addressing the complex, multi-objective challenges inherent in
vehicular networks. This section reviews existing approaches across
three key domains: GAN-based vehicular networks, generative AI
methodologies beyond GANs, and federated learning applications in
IoV environments.

\subsubsection{GAN-based Vehicular Networks}

The HVS-GAN framework~\cite{wu2024federated} represents a notable
advancement in federated GAN training for heterogeneous vehicle
scheduling in IoV scenarios. This approach addresses the
heterogeneity of intelligent connected vehicles (ICVs) by
incorporating Deep Deterministic Policy Gradients (DDPG) learning to
manage varying model size constraints and training costs. However,
the framework primarily focuses on scheduling optimization without
adequately addressing the multi-objective trade-offs between
learning efficiency, network overhead, and latency that are critical
in real-time vehicular applications. Furthermore, the approach
employs a single discriminator architecture, which limits its
ability to simultaneously optimize multiple competing objectives
such as data quality, communication efficiency, and computational
resource utilization.
Liao et al.~\cite{zhao2020novel} proposed a novel GAN-based vehicle
trajectory prediction method that incorporates a spatio-temporal
attention mechanism within the GAN generator to achieve optimal
weight distributions of various spatiotemporal features. While
their approach demonstrates impressive performance, the method
suffers from computational complexity issues that make real-time
deployment challenging in resource-constrained vehicular
environments.

The application of GANs in vehicular environments has also focused
on security and anomaly detection. Seo et al.~\cite{seo2018gids}
introduced GIDS (GAN-based Intrusion Detection System), one of the
first implementations of GANs for in-vehicle network security. Their
approach demonstrated the potential of GANs for generating synthetic
attack patterns to improve intrusion detection in vehicular
networks. However, the framework lacks consideration for resource
efficiency and computational overhead.
More recently, Ioannou and Vassiliou~\cite{ioannou2026gan} present a
comprehensive benchmark of five GAN architectures (including
Wasserstein GAN with gradient penalty) for intrusion detection in
resource-constrained IoT deployments, introducing an
accuracy-per-joule evaluation metric to explicitly quantify the
energy cost of GAN-based detection under severe class imbalance.
This work highlights that architecture choice materially affects
energy efficiency in constrained edge deployments, reinforcing that
resource-awareness must be evaluated jointly with detection accuracy
rather than as an afterthought -- a principle central to the joint
optimization objective proposed in this paper.

\subsubsection{Generative AI based Vehicular Networks}
Beyond GANs, various generative AI methodologies have been explored
for vehicular network applications, including diffusion models,
transformer-based approaches, and variational autoencoders.

DiffusionDrive~\cite{liao2025diffusiondrive} introduces a truncated
diffusion model for end-to-end autonomous driving, achieving a
remarkable 10$\times$ reduction in denoising steps compared to
vanilla diffusion policies. The approach incorporates prior
multi-mode anchors and a cascade diffusion decoder for enhanced
interaction with conditional scene context. However, despite the
computational improvements, the method still requires substantial
computational resources for real-time inference, making it
challenging to deploy in resource-constrained vehicular
environments.
The GAI-IoV framework~\cite{xie2024gai} bridges generative AI and
vehicular networks through edge intelligence, enabling cooperation
between roadside units (RSUs) and vehicles for multi-modal content
generation. While the framework addresses some distributed computing
challenges, it primarily focuses on content generation without
considering the multi-objective optimization requirements of modern
IoV systems.
Conditional VAE networks for vehicle trajectory prediction have
demonstrated better performance with fewer parameters across
different time horizons compared to traditional
approaches~\cite{de2022vehicles}. While computationally more
efficient than some alternatives, these methods still lack the
capability to simultaneously optimize multiple competing objectives
such as prediction accuracy, computational efficiency, and
communication overhead.

\subsubsection{Federated Learning in IoV}
Federated Learning has emerged as a promising paradigm for IoV
applications, enabling collaborative model training while preserving
data privacy and reducing communication overhead~\cite{lyu2026enabling}. However, existing
approaches face significant challenges when dealing with
multi-objective optimization requirements and resource efficiency
constraints.
The multi-objective distributed client selection
framework~\cite{cha2024multi} for federated learning-assisted IoV
addresses the cost of maintaining active states for participating
vehicles. The approach employs a distributed scheme to reduce
communication overhead and improve system efficiency. However, the
method primarily focuses on client selection optimization without
adequately addressing the learning efficiency and model performance
trade-offs.
Lu et al.~\cite{lu2025clustered} represent a significant advancement
in applying FL to 6G-enabled intelligent transportation systems. The
approach demonstrates effective real-time traffic flow management
capabilities. Nevertheless, the framework suffers from scalability
issues when dealing with large-scale vehicular networks, and the
communication overhead increases significantly with the number of
participating vehicles. The single-objective focus on traffic
management neglects other critical IoV requirements such as energy
efficiency, latency optimization, and resource utilization.
Nie et al.~\cite{nie2025mobility} introduce a two-level MEC-assisted
framework with a mobility-supported participant decision algorithm.
While this addresses some mobility challenges, the centralized
aggregation process creates bottlenecks that impact system latency
and scalability, particularly in dense vehicular environments.

These identified gaps underscore the need for a novel approach that
can simultaneously address multiple optimization objectives while
maintaining practical feasibility for real-world IoV deployments.
Our proposed framework addresses these limitations by introducing a
multi-discriminator architecture within a federated learning
paradigm, specifically designed to optimize the trade-offs between
learning efficiency, resource utilization, network overhead, and
latency in vehicular environments.

\section {System Model} \label{systemmodel} 
\begin{figure}[tbp]
\centerline{\includegraphics[width=0.85\linewidth]{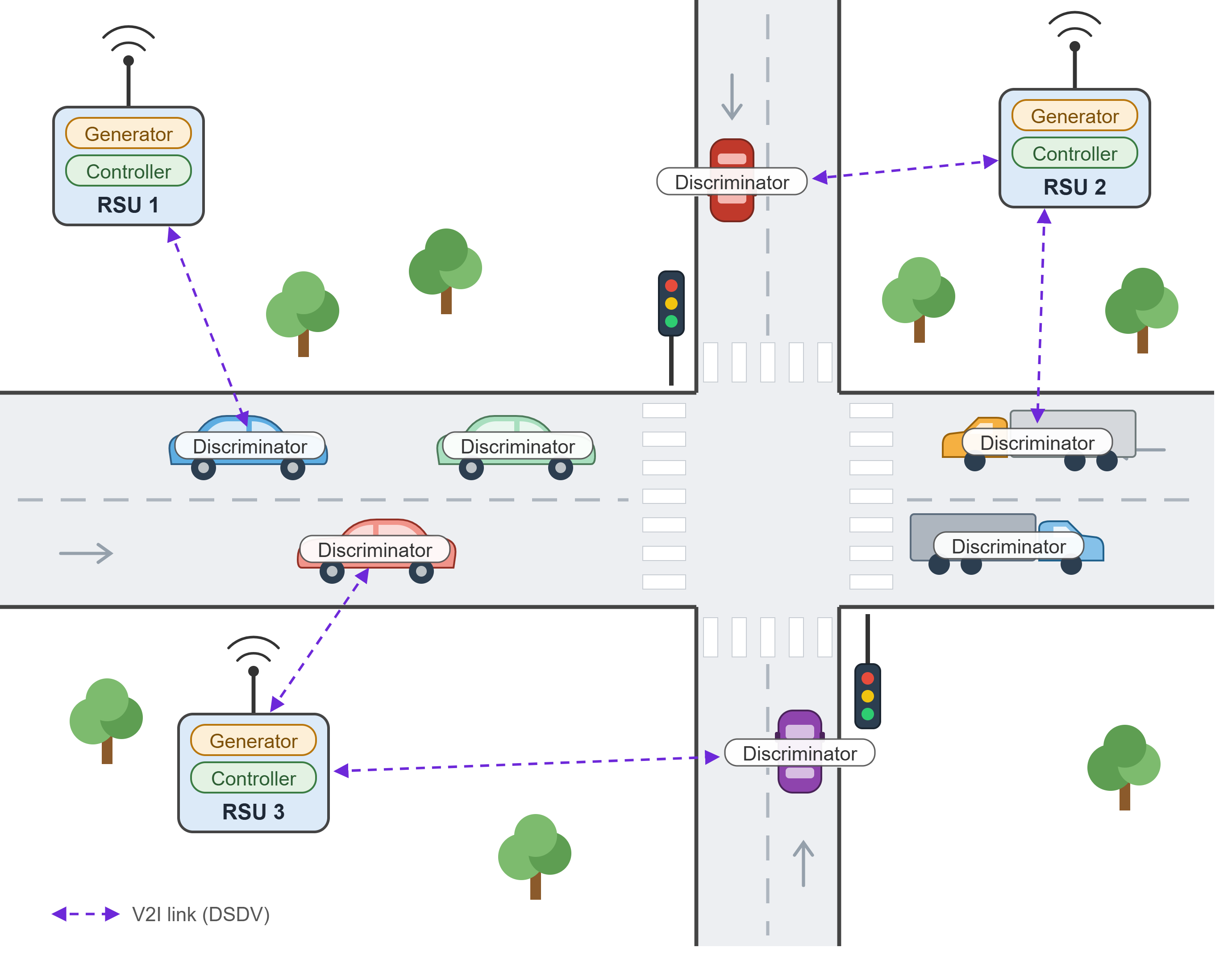}}
\caption{Architectural view of the proposed framework } 
\label{fig111}
\end{figure}

The overall architectural design of the proposed MD-WGAN framework
in IoV environments is illustrated in Fig.~\ref{fig111}. As shown,
each RSU hosts a generator $g_i$ paired with a local controller
(combining reinforcement learning and game-theoretic coordination)
that manages its training, while each vehicle acts as a
discriminator $D_j$ evaluating synthesized data against real traffic
samples over the V2I link. This section describes the system model
underlying this architecture, the network topology, wireless
communication model, and computational resource model, while the
controller's coordination mechanisms are detailed in
Section~\ref{framework}.

\subsection{Network and System Model}
\subsubsection{IoV Network Architecture}
We consider an IoV network consisting of:
\begin{itemize}
\item \textbf{Roadside Units (RSUs)}: $\mathcal{G} = \{1, 2, \ldots, N_G\}$ RSUs host the generators along with RL agents, leveraging their stationary and resource-rich infrastructure for adaptive decision making.

\item \textbf{Vehicles}: $\mathcal{D} = \{1, 2, \ldots, K\}$ vehicles moving along the road act as discriminators, evaluating the quality of generated data against real vehicular samples under mobility constraints.

\item \textbf{V2I Communication}: Communication between RSUs and vehicles occurs through V2I wireless links.
\end{itemize}

As vehicles traverse the road, a discriminator may leave the coverage area of one RSU and enter the range of another. In such cases, the communication link is seamlessly re-established with the new generator, ensuring continuity of training and feedback exchange.

\subsubsection{WGAN Training Model}
In the distributed training environment, each generator $g_i \in \mathcal{G}$ is paired with a subset of discriminators $\mathcal{D}_{g_i} \subseteq \mathcal{D}$. 

\textbf{Generator Operations}: 
Each generator $g_i$ takes an observed trajectory context $c$ (a window of past
feature vectors) together with a latent variable $z \sim p_z$, and predicts the
next-step feature vector $\hat{y} = G(c, z)$. Training pairs $(c, y)$ are drawn
from the dataset sequences, where $y$ is the ground-truth continuation of $c$.

\textbf{Discriminator Operations}: Each discriminator $D_j$ evaluates both real samples $x \sim p_{\text{data}}$ and generated samples $\tilde{x} \sim p_g$ to distinguish between real and synthetic data.

\textbf{Wasserstein Distance}: The training objective follows the Wasserstein GAN formulation:
\begin{small}
\begin{equation} 
\begin{aligned} \label{eq:lossgen}
\mathcal{L}_{WGAN}^{(t)}(G, D_j) = &\mathbb{E}_{x \sim P_r^{(t)}}[D_j^{(t)}(x)] - \mathbb{E}_{\tilde{x} \sim P_g^{(t)}}[D_j^{(t)}(\tilde{x})] - \\
&\lambda GP(D_j^{(t)})
\end{aligned}
\end{equation}
\end{small}
where $GP(D_j^{(t)})$ is the gradient penalty term ensuring Lipschitz continuity, $\mathbb{E}_{x\sim P_r^{(t)}}[\cdot]$ denotes expectation with respect to the
real-data distribution , and $\mathbb{E}_{\tilde x\sim P_g^{(t)}}[\cdot]$ is expectation over the
model (generated) distribution induced by the generator at round $t$.
For clarity, Table \ref{tab:variables} summarizes the main symbols and variables used throughout the paper, including system parameters, optimization variables, and reinforcement-learning terms.
\begin{table}[t]
\centering
\scriptsize
\setlength{\tabcolsep}{4pt}
\renewcommand{\arraystretch}{1.05}
\caption{List of Symbols and Variables}
\label{tab:variables}
\begin{tabular}{ll}
\toprule
\textbf{Symbol} & \textbf{Description} \\
\midrule
$G=\{1,\dots,N_G\}$ & Set of generator nodes (RSUs). \\
$D=\{1,\dots,K\}$ & Set of discriminators (vehicles). \\
$V=G\cup D$ & All nodes in the system. \\
$g_i, D_j$ & Indices for generator and discriminator. \\
$D_{g_i}$ & Discriminator subset assigned to $g_i$. \\
$z\!\sim\!p_z$, $x\!\sim\!P_r$, $\tilde{x}\!\sim\!P_g$ & Latent, real, and generated samples. \\
$c$ & Observed trajectory context (window of $W$ past feature vectors). \\
$G(c,z)$ & Generator output (synthetic data). \\
$D_j(\cdot)$ & Output of discriminator $j$. \\
$L_{\mathrm{WGAN}}(G,D_j)$ & Wasserstein loss with gradient penalty. \\
$\lambda_{\mathrm{GP}}$ & Gradient penalty coefficient. \\
$h_{i,j}$ & Channel gain between $g_i$ and $d_j$. \\
$d_{i,j}$, $\beta$ & Distance and path-loss exponent. \\
$R_{i,j}$ & Communication rate. \\
$B_n$ & Bandwidth (Hz). \\
$L_{D_{g_i}}, T_{D_{g_i}}$ & Avg. latency and network traffic of $g_i$. \\
$C_{D_{g_i}}, M_{D_{g_i}}$ & CPU and memory usage metrics. \\
$\alpha_G, \alpha_{1:4}$ & Weighting coefficients in the objective. \\
$E_{\text{epochs}}$ & Number of training epochs. \\
$C_{\max}, M_{\max}$ & Max CPU and memory capacities. \\
$s_t, a_t$ & RL state and action at time $t$. \\
$c^{(t)}_{g_i}$ & Cost for generator $g_i$. \\
$Q(s,a)$ & Action–value function (DQN). \\
$\epsilon, \gamma, \alpha$ & Exploration, discount, learning rates. \\
$\vartheta_{g_i}$ & RL-selected config vector for $g_i$
\\[-0.2em] & (e.g., learning rate, batch size, sync interval) \\
$U_D, U_G$ & Utility functions in the coordination game. \\
$E_D, E_G$ & Epoch allocations (strategies). \\
$D_{\text{ens}}$ & Ensemble discriminator $\frac{1}{|S|}\sum_{i\in S} w_i D_j$. \\
$d, B, S_{\text{sync}}$ & Model parameters, batch size, sync steps,  \\
$K_S$& active discriminators.\\
\bottomrule
\end{tabular}
\vspace{-0.4em}
\end{table}

\subsection{Communication Model}

\subsubsection{Wireless Channel Model}
The communication between generator $g_i$ and discriminator $D_j$
experiences channel gain $h_{i,j}$ following the log-distance path
loss model~\cite{rappaport2002wireless}:
\begin{equation}
h_{i,j} = h_0 \left(\frac{d_0}{d_{i,j}}\right)^\beta
\end{equation}
where $d_{i,j}$ is the distance between $g_i$ and $D_j$, $\beta$ is
the path loss exponent, and $h_0$ is the reference channel gain at
distance $d_0$.

\subsubsection{Data Transmission}
The data transmission rate from discriminator $D_j$ to generator
$g_i$ is given by the Shannon-Hartley channel capacity
formula~\cite{rappaport2002wireless}:
\begin{equation}
R_{i,j} = B_n \log_2\left(1 + \frac{P_{tx} h_{i,j}}{N_0 B_n}\right)
\end{equation}
where $B_n$ is the channel bandwidth, $P_{tx}$ is the transmission
power, and $N_0$ is the noise power spectral density.

The communication latency and network traffic overhead are:
\begin{align}
L_{D_{g_i}}^{(t)} &= \frac{1}{|\mathcal{D}_{g_i}|}\sum_{d \in \mathcal{D}_{g_i}} Latency_{g_i,d}(t) \\
T_{D_{g_i}}^{(t)} &= \frac{1}{|\mathcal{D}_{g_i}|}\sum_{d \in \mathcal{D}_{g_i}} Traffic_{g_i,d}(t)
\end{align}
We emphasize that this latency characterizes the round-trip time
of a federated training/synchronization cycle (covering sample
broadcast, discriminator evaluation, feedback aggregation, and
generator update) and not the inference-time latency of the
trained generator when deployed for a downstream driving task.
\subsection{Computation Model}

\subsubsection{Resource Allocation}
Each generator $g_i$ allocates computing resources (CPU and memory) to associated discriminators. Similarly, each vehicular node has computational and memory capacity constraints.

\subsubsection{Resource Utilization}
CPU and memory utilization affects performance.
\begin{align}
C_{D{g_i}}^{(t)} &= \frac{1}{|\mathcal{D}_{g_i}|}\sum_{d \in \mathcal{D}_{g_i}} CPU_d^{used}(t) \\
M_{D{g_i}}^{(t)} &= \frac{1}{|\mathcal{D}_{g_i}|}\sum_{d \in \mathcal{D}_{g_i}} MEM_d^{used}(t)
\end{align}

\section{Adaptive Multi-Discriminator Framework}
\label{framework}

The proposed MD-WGAN framework coordinates generator and
discriminator training through four interacting mechanisms, each
addressing a distinct question in the training loop: \emph{what}
should be optimized at each round (Section~\ref{Objective Optimization}),
\emph{who} decides which discriminators participate and how training
epochs are divided between generators and discriminators
(Sections~\ref{sec:rl_engine} and~\ref{sec:nash_coordination}), and
\emph{how} the resulting decisions are executed and monitored across
distributed vehicular nodes (Sections~\ref{sec:distributed_training}
and~\ref{sec:monitoring}). Section~\ref{sec:overview} first presents
the adaptation cycle that ties these four mechanisms together into a
single recurring loop, before each mechanism is detailed in turn.

\subsection{Overview: The Adaptation Cycle}
\label{sec:overview}
\begin{figure*}[tbp]
\centerline{\includegraphics[width=0.8\linewidth]{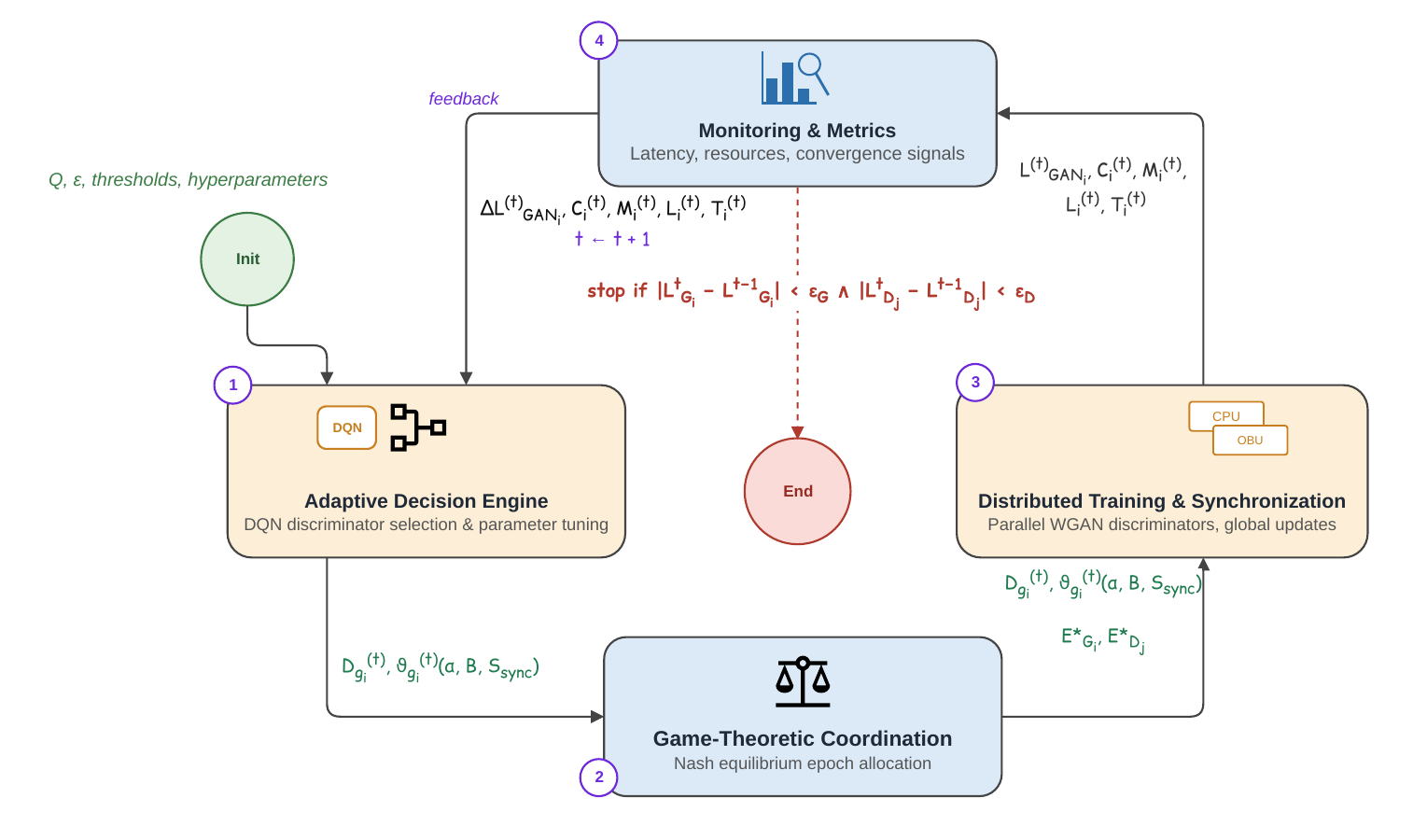}}
\caption{Workflow of proposed approach methodology} 
\label{fig1}
\end{figure*}

Our MD-WGAN framework addresses the dynamic challenges of IoV environments through a four-step cyclical methodology that continuously adapts to changing network conditions. The framework integrates reinforcement learning for local optimization with game-theoretic coordination for global resource management, ensuring optimal performance across distributed vehicular nodes
(Figure~\ref{fig1}).

To clearly illustrate the continuous adaptation cycle, the overall workflow of Steps 1–4 is expressed as a recurrent loop. At each iteration, the system first executes the Adaptive Decision Engine (Step 1) to select discriminators and tune parameters, then applies Game-Theoretic Coordination (Step 2) to allocate training epochs fairly across nodes. Next, the framework performs Distributed Training and Synchronization (Step 3) using the optimized configurations. Finally, the system carries out Monitoring and Metrics collection (Step 4), which provides updated feedback for the next cycle. This loop repeats continuously until the convergence criteria are satisfied (Algorithm~1).

\begin{algorithm}[t]\small
\caption{Continuous Adaptation Cycle of the MD\textnormal{-}WGAN for each generator and its related discriminators}
\label{alg:mdwgan-cycle}
\begin{algorithmic}[1]
\Require Initial $\theta_G,\theta_D$, Q-network $Q$, thresholds $\epsilon_G,\epsilon_D$
\Ensure Optimized generator/discriminator parameters
\State $t \gets 0$; initialize replay buffer $\mathcal{B}$ and hyperparameters
\Repeat
    \State \textbf{Step 1: Adaptive Decision Engine}
    \State $a_t \gets \pi_\theta(s_t)$ \Comment{$\epsilon$-greedy from DQN policy}
    \State $Q(s_t,a_t) \gets Q(s_t,a_t) + \alpha\!\left[r_t + \gamma \max_{a} Q(s_{t+1},a) - Q(s_t,a_t)\right]$
    \State \textbf{Step 2: Game-Theoretic Coordination}
    \State $(E_D^{*}, E_G^{*}) \gets \textsc{BestResponse}\big(U_D(\cdot), U_G(\cdot)\big)$
    \State \textbf{Step 3: Distributed Training \& Synchronization}
    \State $\theta_G \gets \theta_G - \eta \nabla_{\theta_G} \mathcal{L}_G$ \Comment{WGAN update w/ GP}
    \State $\theta_D \gets \theta_D - \eta \nabla_{\theta_D} \mathcal{L}_D$
    \State \textbf{Step 4: Monitoring \& Metrics}
    \State $s_{t+1} \gets f(\text{lat},\text{cpu},\text{mem},\text{traf},\text{tloss})$
    \Statex \hspace{\algorithmicindent} \Comment{abbreviations: latency, CPU\_usage, memory\_usage, network\_traffic\_overhead, training\_loss}
    \State $t \gets t + 1$
\Until{$|\mathcal{L}_G^{t}-\mathcal{L}_G^{t-1}| < \epsilon_G$ \textbf{and} $|\mathcal{L}_D^{t}-\mathcal{L}_D^{t-1}| < \epsilon_D$}
\end{algorithmic}
\end{algorithm}

The framework begins by establishing the distributed training environment, where RSUs function as generators and vehicles operate as discriminators. This configuration leverages the stationary, resource-rich RSUs for data generation while utilizing the mobile vehicles to provide diverse, context-specific discrimination feedback across wireless communications.
Initial hyperparameters, thresholds, and baseline metrics are defined. Each generator node is paired with a reinforcement learning agent, initialized with Q-tables or policy networks to support adaptive decision-making during training.
Once the initial setup is complete, the framework operates through a series of coordinated steps that enable dynamic optimization of the federated GAN training process.

\subsection{Joint Optimization Objective}
\label{Objective Optimization}
Based on the system model described above, we formulate the joint discriminator selection and resource allocation problem. The objective is to minimize the Wasserstein distance while satisfying resource and network constraints in the IoV environment.

For each generator $g_i \in \mathcal{G}$ and its associated discriminator subset $\mathcal{D}_{g_i} \subseteq \mathcal{D}$, the optimization problem is formulated as:

\begin{equation}
\begin{aligned}\label{eq:md-wgan-formulation}
\min_{\boldsymbol{\theta}_G, \{\boldsymbol{\theta}_{D_j}\}_{j=1}^K, \mathcal{D}_{g_i}} 
& \alpha_G \mathcal{L}_{WGAN}^{(t)}(G, \mathcal{D}_{g_i}) + \alpha_1 \cdot \frac{C_{D_{g_i}}^{(t)}}{C_{max}} \\
& + \alpha_2 \cdot \frac{M_{D_{g_i}}^{(t)}}{M_{max}} + \alpha_3 \cdot \frac{L_{D_{g_i}}^{(t)}}{L_{max}} + \alpha_4 \cdot \frac{T_{D_{g_i}}^{(t)}}{T_{max}}
\end{aligned}
\end{equation}

subject to:
\begin{align}
& C_{D_{g_i}}^{(t)} \leq C_{max}, && \forall g_i \in \mathcal{G}, D_{g_i} \in \mathcal{D}\label{eq:constraint_cpu} \\
& M_{D_{g_i}}^{(t)} \leq M_{max}, && \forall g_i \in \mathcal{G}, D_{g_i} \in \mathcal{D}\label{eq:constraint_mem} \\
& L_{D_{g_i}}^{(t)} \leq L_{max}, && \forall g_i \in \mathcal{G}, D_{g_i} \in \mathcal{D} \label{eq:constraint_latency} \\
& T_{D_{g_i}}^{(t)} \leq T_{max}, && \forall g_i \in \mathcal{G}, D_{g_i} \in \mathcal{D} \label{eq:constraint_traffic} \\
& \big| Loss_v^{t+1} - Loss_v^t \big| \leq \epsilon, && \forall v \in \mathcal{G} \cup \mathcal{D} \label{eq:constraint_convergence}
\end{align}

where $\mathcal{V} = \mathcal{G} \cup \mathcal{D}$ represents all vehicular nodes in the system and
each coefficient in Eq.~(\ref{eq:md-wgan-formulation}) acts as a priority
weight that determines how strongly the corresponding term
influences discriminator selection and resource allocation.
Because $\sum_i \alpha_i = 1$, the coefficients define a convex
combination over five competing objectives (training effectiveness
and four resource/network costs) so increasing any one
coefficient necessarily reduces the relative influence of the
others.

\subsubsection{Objective Function Components}

\begin{itemize}
\item \textbf{WGAN Training Loss} ($\alpha_G \mathcal{L}_{WGAN}^{(t)}$): Measures the Wasserstein distance between real and generated data distributions as defined in Equation~(\ref{eq:lossgen}).

\item \textbf{CPU Utilization} ($\alpha_1 \cdot \frac{C_{D_{g_i}}^{(t)}}{C_{max}}$): Average normalized CPU usage across selected discriminators. This term penalizes high processor utilization to prevent resource exhaustion. Also, $C_{max} = 100$ represents full CPU capacity per vehicle.

\item \textbf{Memory Consumption} ($\alpha_2 \cdot \frac{M_{D_{g_i}}^{(t)}}{M_{max}}$): Average normalized memory usage across selected discriminators. This ensures sufficient memory remains available for other tasks. $M_{max} = \max_{v \in \mathcal{V}} MEM_v^{capacity}$ is the maximum memory capacity across all nodes. 

\item \textbf{Communication Latency} ($\alpha_3 \cdot \frac{L_{D_{g_i}}^{(t)}}{L_{max}}$): Average normalized communication delay between generator $g_i$ and its discriminators. $L_{max}$ is the maximum acceptable latency. Lower values indicate faster feedback exchange.

\item \textbf{Network Traffic Overhead} ($\alpha_4 \cdot \frac{T_{D_{g_i}}^{(t)}}{T_{max}}$): Average normalized data transmission volume. $T_{max}$ represents the maximum bandwidth for bidirectional communication. This term encourages efficient communication protocols and reduces network congestion.

\end{itemize}


\subsubsection{Decision Variables}

The optimization involves determining:
\begin{itemize}
\item \textbf{Model Parameters}: Generator parameters $\boldsymbol{\theta}_G$ and discriminator parameters $\{\boldsymbol{\theta}_{D_j}\}_{j=1}^K$ (trainable neural network weights)
\item \textbf{Discriminator Selection}: Subset $\mathcal{D}_{g_i} \subseteq \mathcal{D}$ for each generator $g_i$
\item \textbf{Resource Allocation}: Computing resources allocated to each discriminator
\end{itemize}

\subsection{Step 1: Adaptive Decision Engine}
\label{sec:rl_engine}

The Adaptive Decision Engine is the intelligent core of the
framework: each generator node maintains an independent Deep
Q-Network (DQN) agent that learns a policy for discriminator subset
selection from environmental feedback.

\subsubsection{RL Formulation}

\begin{itemize}
\item \textbf{State Space}: $\mathcal{S} \in \mathbb{R}^7$ -- a
7-dimensional vector $s_t$ aggregating, for generator $g_i$ and its
associated discriminators, the average discriminator loss, average
discriminator score, communication latency, CPU utilization, memory
consumption, network traffic, and the discriminator-side average
score from the previous round:
\begin{equation}
\begin{gathered}
s_t = [\bar{\mathcal{L}}_D,\ \bar{s}_D,\ latency,\ cpu\_usage,
memory\_usage,\ \\traffic,\ \bar{s}_D^{(t-1)}]
\end{gathered}
\end{equation}

\item \textbf{Action Space}: $\mathcal{A} = \{a_1, \ldots,
a_{|\mathcal{A}|}\}$, where $|\mathcal{A}| = K_{\text{local}}$, the
number of discriminators in the local pool assigned to generator
$g_i$. The DQN outputs one Q-value per candidate discriminator in
this pool; the top-$k$ discriminators by Q-value are selected as the
active subset $\mathcal{D}_{g_i}^{(t)}$ for round $t$. This
per-discriminator scoring scheme avoids the combinatorial blowup of
enumerating all $\binom{K_{\text{local}}}{k}$ subsets directly, at
the cost of not modeling interactions between simultaneously
selected discriminators.

\item \textbf{Network Architecture}: Fully connected layers
$[128, 128, |\mathcal{A}|]$ with ReLU activations.

\item \textbf{Learning Algorithm}: DQN with experience replay buffer
$\mathcal{B}$ (capacity 10{,}000).
\end{itemize}

The Q-learning update rule follows:
\begin{equation}
\begin{gathered}
Q(s_t,a_t) \leftarrow Q(s_t,a_t) + \alpha \Bigl[ r^{(t)}_{g_i} 
+\, \gamma \max_{a\in\mathcal{A}} Q(s_{t+1},a) - \\Q(s_t,a_t) \Bigr]
\end{gathered}
\label{eq:q_update}
\end{equation}
with actions selected greedily as $a_t \in \arg\max_{a\in\mathcal{A}}
Q(s_t,a)$ during exploitation.

\subsubsection{Bootstrapping and Discriminator Selection}

At initialization, the decision engine selects discriminators in
three phases: (1) an $\varepsilon$-greedy \emph{random exploration
phase} ($\varepsilon_0 = 1.0$) samples broadly across all available
discriminators $\{D_1,\ldots,D_K\}$ to gather diverse experience; (2)
a \emph{performance-based filtering} phase ranks discriminators by
convergence speed and loss reduction, $rank(D_j) = w_1 \cdot
\text{conv\_rate}_i + w_2 \cdot \Delta\mathcal{L}_i$; and (3)
\emph{adaptive subset formation} selects the top-$k$ discriminators
by this ranking as the active subset for training.

\subsubsection{Reward Formulation}
\label{sec:reward_formulation}

To align the RL objective with the global optimization in
Eq.~(\ref{eq:md-wgan-formulation}), we define a per-round reward for generator
$g_i$ that balances training effectiveness against resource and
communication cost:
\begin{equation}
r_{g_i}^{(t)} = w_1 \frac{\Delta\mathcal{L}_{WGAN_{g_i}}^{(t)}}
{\mathcal{L}_{WGAN_{g_i}}^{(t-1)}}
- w_2 \hat{L}_{D_{g_i}}^{(t)} - w_3 \hat{T}_{D_{g_i}}^{(t)}
- w_4 \hat{C}_{D_{g_i}}^{(t)} - w_5 \hat{M}_{D_{g_i}}^{(t)}
\label{eq:reward}
\end{equation}
where $\Delta\mathcal{L}_{WGAN_{g_i}}^{(t)} =
\mathcal{L}_{WGAN_{g_i}}^{(t-1)} - \mathcal{L}_{WGAN_{g_i}}^{(t)}$ is
the relative improvement in GAN loss, $\hat{L}, \hat{T}, \hat{C},
\hat{M}$ denote latency, network traffic, CPU, and memory usage
normalized to $[0,1]$, and $(w_1,\ldots,w_5)$ are reward
weights satisfying $\sum_i w_i = 1$ (Section~\ref{sec:app-specific}
discusses application-specific weight configurations). The agent
maximizes the expected discounted reward via
Eq.~(\ref{eq:q_update}).

\subsubsection{Exploration Schedule}

Exploration decreases over training as
\begin{equation}
\varepsilon_t = \max(\varepsilon_{\min}, \varepsilon_0 \cdot
\varepsilon_{\text{decay}}^t)
\end{equation}
maintaining a minimum exploration rate to remain responsive to
network dynamics (values in Table~\ref{tab:rl_hyperparams}).

\subsubsection{Decision Output}

At each round, the decision engine outputs the selected
discriminator subset $\mathcal{D}_{g_i}^{(t)}$ used by the
distributed training step (Section~\ref{sec:distributed_training}).
Extending the decision output to jointly select per-round training
configurations (learning rate, batch size, synchronization interval)
is left for future work; the current implementation fixes these
hyperparameters (Tables~\ref{tab:rl_hyperparams},\ref{tab:nash_hyperparams}).

\subsection{Step 2: Game-Theoretic Best-Response Coordination}
\label{sec:nash_coordination}

The framework models the interaction between discriminators and
generators as a strategic game: each discriminator (player) seeks to
maximize its own training effectiveness while accounting for the
generator's competing actions, and vice versa. We formulate the
training epoch allocation problem between discriminators and
generators as a game and solve it via a bounded best-response search
over discretized strategy sets, described in
Section~\ref{sec:brs_procedure}.

\subsubsection{Utility Functions}

For a strategy combination $(E_D, E_G)$, where $E_D$ and $E_G$
denote epoch allocations for the discriminator and generator
respectively, the utilities balance three components: (i) the
\emph{training effectiveness} gained from allocating epochs, (ii)
the \emph{resource cost} of excessive computational or communication
consumption, and (iii) the \emph{interference term}, reflecting
performance degradation caused by the competing player's actions:
\begin{align}
U_D(E_D, E_G) &= f_D(E_D) - c_D(E_D) - I_G(E_G) \label{eq:utility_d}\\
U_G(E_D, E_G) &= f_G(E_G) - c_G(E_G) - I_D(E_D) \label{eq:utility_g}
\end{align}
where $f_D, f_G$ are training effectiveness gains, $c_D, c_G$ are
computational resource costs, and $I_G, I_D$ are interference terms
from the opponent's actions.

\subsubsection{Component Instantiation}

Training effectiveness is modeled as a diminishing-returns function
of allocated epochs, reflecting that additional epochs yield
progressively smaller convergence gains:
\begin{equation}
f_D(E_D) = 1 - e^{-\kappa_D E_D}, \quad
f_G(E_G) = 1 - e^{-\kappa_G E_G}
\end{equation}

Resource cost is linear in the number of epochs, scaled relative to
the effectiveness term:
\begin{equation}
\begin{gathered}
c_D(E_D) = \lambda_D \cdot c^{\text{epoch}}_D \cdot E_D, \\
c_G(E_G) = \lambda_G \cdot c^{\text{epoch}}_G \cdot E_G
\end{gathered}
\end{equation}
where $c^{\text{epoch}}_D, c^{\text{epoch}}_G$ denote the per-epoch
computational cost of a training round for the discriminator and
generator, respectively, and $\lambda_D, \lambda_G$ scale this cost
relative to effectiveness.

Interference terms are modeled as proportional to the opponent's own
effectiveness, capturing that one player's high resource consumption
degrades the shared training environment for the other:
\begin{equation}
I_G(E_G) = \gamma_{\text{nash}} \cdot f_G(E_G), \quad
I_D(E_D) = \gamma_{\text{nash}} \cdot f_D(E_D)
\end{equation}

The parameters $\kappa_D, \kappa_G, c^{\text{epoch}}_D,
c^{\text{epoch}}_G, \lambda_D, \lambda_G, \gamma_{\text{nash}}$ are
given in Table~\ref{tab:nash_hyperparams}. These values were set
based on the observed per-epoch computational cost and convergence
behavior during the exploration phase of training.

\subsubsection{Best-Response Search Procedure}
\label{sec:brs_procedure}

Given discrete strategy sets $\mathcal{E}_D$ and $\mathcal{E}_G$, the
coordination step computes an approximate equilibrium via a two-pass
best-response search: for each candidate $E_D \in \mathcal{E}_D$,
the best response $E_G$ maximizing $U_G$ is found, yielding a
candidate pair $(E_D^{(1)}, E_G^{(1)})$; symmetrically, for each
candidate $E_G \in \mathcal{E}_G$, the best response $E_D$
maximizing $U_D$ is found, yielding $(E_D^{(2)}, E_G^{(2)})$. The
pair with the higher joint payoff $U_D + U_G$ is selected as
$(E_D^*, E_G^*)$ for that round. Because this procedure evaluates
best responses over discretized strategy sets in a bounded number of
passes rather than iterating to a fixed point, we refer to it as
\emph{best-response coordination} rather than a converged Nash
equilibrium.

\subsubsection{Effect on Coordination}

This coordination step improves resource allocation across nodes,
provides a principled epoch distribution between generator and
discriminators in place of a fixed schedule, and reduces conflicting
parameter updates that would otherwise arise from uncoordinated
independent training. 

\subsection{Step 3: Distributed Training and Synchronization}
\label{sec:distributed_training}

Using the discriminator subset from Step 1 and the training epoch
allocation from Step 2, the framework executes distributed WGAN
training across vehicular nodes at round $t$. The system employs $K$
discriminators $\{D_1, D_2, \ldots, D_K\}$ operating in parallel,
each observing a distinct local data partition.

\subsubsection{Generator Training via Ensemble Feedback}

Each generator $g_i$ receives feedback from an ensemble of its
selected discriminators:
\begin{equation}
\mathcal{L}_{g_i}^{\mathrm{WGAN}} = -\, \mathbb{E}_{z\sim p_z}
\big[\, D_{\mathrm{ens}}(G(c,z)) \,\big]
\end{equation}
where the ensemble discriminator is
\begin{equation}
D_{\text{ens}} = \frac{1}{|S|} \sum_{i \in S} w_i \cdot D_j,
\end{equation}
$S \subseteq \{1, \ldots, K\}$ is the dynamically selected
discriminator subset, $z \sim p_z$ is a latent vector drawn from
prior $\mathcal{Z}$, and $G: \mathcal{Z} \to \mathcal{X}$ is the
generator, so $G(z)$ is a synthetic sample.

\subsubsection{Discriminator Training}

Each discriminator $D_j$ is trained independently using the
Wasserstein loss with gradient penalty:
\begin{equation}
\begin{aligned}
\mathcal{L}_{D_j} =\ &
\mathbb{E}_{x \sim p_{\text{data}}}[D_j(x)]
- \mathbb{E}_{z \sim p_z}[D_j(G(c,z))] \\
&+\, \lambda_{GP}\, \mathbb{E}_{\hat{x} \sim p_{\hat{x}}}
\left[ \left(\lVert \nabla_{\hat{x}} D_j(\hat{x}) \rVert_2 - 1
\right)^2 \right]
\end{aligned}
\label{eq:disc_loss}
\end{equation}
where $x \sim p_{\text{data}}$ is a real sample from the data domain
$\mathcal{X}$, $\|\nabla_{\hat{x}} D_j(\hat{x})\|_2$ is the gradient
of $D_j$ with respect to its input $\hat{x}$, and $\lambda_{GP} > 0$
is the WGAN-GP coefficient.

\subsection{Step 4: Monitoring and Convergence Criteria}
\label{sec:monitoring}

The monitoring system continuously collects performance and resource
metrics to inform the next adaptation cycle, spanning three
dimensions:

\begin{itemize}
\item \textbf{Network performance}: communication latency between
vehicular nodes, bandwidth utilization and communication overhead,
and network topology changes and connection stability;
\item \textbf{Resource utilization}: CPU usage patterns across
participating vehicles, and memory consumption trends and peak
usage events;
\item \textbf{Training quality}: convergence rate for generators and
discriminators, loss function evolution and stability, and training
efficiency (iterations per unit time).
\end{itemize}

These metrics feed into a multi-level convergence criterion,
requiring both generator and discriminator losses to stabilize
before the training cycle terminates:
\begin{equation}
C_{\text{loss}} = \bigl(|L_G^{(t)} - L_G^{(t-1)}| < \varepsilon_G\bigr)
\land \bigl(|L_D^{(t)} - L_D^{(t-1)}| < \varepsilon_D\bigr)
\label{eq:loss}
\end{equation}
Monitoring results feed back into the state vector $s_t$
(Section~\ref{sec:rl_engine}), closing the adaptation loop.

\section{Experiments and Results} \label{simulation}

\subsection{Experiment Setup}

To evaluate the effectiveness of the proposed adaptive MD-WGAN framework, we implemented a comprehensive simulation environment integrating NS3 network simulator with PyTorch-based adaptive GAN training pipeline. The vehicular network is modeled using NS3's 802.11p stack to simulate realistic communication patterns. Key simulation parameters are summarized in Table~\ref{tab:sim_params}.
the RL agent and game-theoretic coordination
hyperparameters used to obtain all results are given in
Tables~\ref{tab:rl_hyperparams}--\ref{tab:nash_hyperparams}. Unless
otherwise noted, these experiments use the reward weighting
$(w_1,\ldots,w_5) = (0.2, 0.2, 0.2, 0.2, 0.2)$; the four
application-specific weightings are introduced separately in
Section~\ref{sec:app-specific}.

The simulation was executed on an Amazon EC2 \texttt{m5.4xlarge} instance equipped with 16 vCPUs and 64~GB of RAM, providing sufficient computational capacity for concurrent NS3 simulation, multi-agent reinforcement learning, and distributed GAN training with real-time parameter adaptation.
The simulation architecture includes multiple RSU nodes functioning as generator agents with embedded RL controllers, and a dynamic fleet of mobile vehicles serving as discriminator nodes. 
Vehicles follow realistic urban traffic patterns generated using SUMO (Simulation of Urban Mobility) traffic simulator, ensuring authentic mobility dynamics. Each curve shows the mean over five independent runs with different random seeds.

\begin{table}[t]
\centering
\caption{Simulation Parameters} 
\label{tab:sim_params}
\begin{tabular}{|l|l|}
\hline
\textbf{Parameter} & \textbf{Value} \\
\hline
Simulation Area & 1500 m $\times$ 1500 m \\
Simulation Time & up to 150 seconds \\
Vehicle Speed Range & 15--60 m/s \\
Mobility Model & SUMO-based Urban Traffic \\
Communication Protocol & DSRC \\
Wireless Standard & IEEE 802.11p \\
Packet Size & 512--2048 bits (adaptive) \\
Number of Generator Nodes & 2 to 8 (distributed RSUs) \\
Number of Discriminator Nodes & 8 to 40 (mobile vehicles) \\
Background Traffic & 150 vehicles \\
RL Learning Rate & 0.001 \\
GAN Learning Rate & 0.0002 (adaptive) \\
Local Training Epochs & 20--50 (adaptive) \\
Batch Size & 32--128 (adaptive) \\
\hline
\end{tabular}
\end{table}
\begin{table}[t]
\centering
\caption{RL Agent (DQN) Hyperparameters}
\label{tab:rl_hyperparams}
\begin{tabular}{ll}
\hline
\textbf{Parameter} & \textbf{Value} \\
\hline
State space dimension $|\mathcal{S}|$ & 7 \\
Network architecture & FC $[128, 128, |\mathcal{A}|]$, ReLU \\
Discount factor $\gamma$ & 0.99 \\
Initial exploration $\epsilon_0$ & 1.0 \\
Minimum exploration $\epsilon_{\min}$ & 0.01 \\
Exploration decay $\epsilon_{\text{decay}}$ & 0.995 \\
Replay buffer capacity $|\mathcal{B}|$ & 10{,}000 \\
Minibatch size & 64 \\
Optimizer & Adam ($lr = 1\times10^{-3}$) \\
Loss function & MSE \\
\hline
\end{tabular}
\end{table}
\begin{table}[t]
\centering
\caption{Game-Theoretic Coordination Parameters}
\label{tab:nash_hyperparams}
\begin{tabular}{ll}
\hline
\textbf{Parameter} & \textbf{Value} \\
\hline
Discriminator effectiveness rate $\kappa_D$ & 0.05 \\
Generator effectiveness rate $\kappa_G$ & 0.10 \\
Per-epoch cost $c_D$ & 0.3 \\
Per-epoch cost $c_G$ & 0.8 \\
Discriminator cost weight $\lambda_D$ & 0.05 \\
Generator cost weight $\lambda_G$ & 0.05 \\
Interference coefficient $\gamma_{\text{nash}}$ & 0.7 \\
Discriminator epoch range $E_D$ & $\{45, 50, \ldots, 75\}$ \\
Generator epoch range $E_G$ & $\{2, 3, \ldots, 15\}$ \\
Solver & Exhaustive best-response search \\
\hline
\end{tabular}
\end{table}

\subsection{Dataset}

The training and evaluation utilize an enhanced version of the NGSIM dataset\footnote{https://data.transportation.gov/stories/s/Next-Generation-Simulation-NGSIM-Open-Data/i5zb-xe34/}, augmented with synthetic vehicular communication data to reflect modern V2X scenarios. The dataset includes comprehensive vehicular mobility and traffic patterns. The dataset was preprocessed into structured tabular format optimized for multi-discriminator GAN training under non-IID conditions.
Each sample corresponds to a $d$-dimensional feature vector
extracted from the NGSIM trajectory data at a fixed sampling
interval. Specifically, each
vector consists of the following features: longitudinal speed,
longitudinal acceleration, lane identifier, headway distance to
lead vehicle, lateral position. Features are computed per vehicle
per timestep from the raw NGSIM trajectory records.
Samples are organized into (context, target) pairs: each context $c$ is a window
of consecutive past feature vectors and the target $y$ is the
subsequent vector. Pairs are split into training and held-out test sets with no
sequence overlap, so RMSE and MAE are computed on unseen continuations.

To simulate realistic distributed scenarios, vehicle trajectories and network data were spatially partitioned such that each discriminator node receives locally relevant, non-overlapping data segments. This partitioning strategy reflects the inherent data heterogeneity in vehicular networks where different geographical regions exhibit distinct traffic patterns and communication characteristics.
Data integration is achieved through Python-based pipelines with real-time preprocessing capabilities linked to each discriminator node.
\subsection{Evaluation Metrics}

The system performance is assessed using comprehensive metrics addressing both training efficiency and performance as below. Accuracy is measured on the held-out test set as the deviation between each
predicted next-step vector $\hat{y}_i = G(c_i, z)$ and its ground-truth
continuation $y_i$.

\begin{itemize}
\item \textbf{Root Mean Square Error (RMSE):} Evaluates prediction accuracy of generated samples compared to ground truth vehicular data:
    \begin{equation}
    \text{RMSE} = \sqrt{\frac{1}{n} \sum_{i=1}^{n} (y_i^{\text{real}} - y_i^{\text{gen}})^2}
    \end{equation}
    
    \item \textbf{Mean Absolute Error (MAE):} Measures average absolute deviation between generated and actual values:
    \begin{equation}
    \text{MAE} = \frac{1}{n} \sum_{i=1}^{n} |y_i^{\text{real}} - y_i^{\text{gen}}|
    \end{equation}
    
    \item \textbf{Latency (seconds):} Measures the time elapsed between the generator’s broadcast of synthetic samples and
    the completion of its update after receiving feedback.
    
    \item \textbf{Network Traffic Overhead (Bytes):} Quantifies the total network overhead over time, including parameter synchronization, state updates, and coordination messages.

   \item \textbf{CPU Usage (\%):} Percentage of computational processing power utilized across all nodes, where each CPU core is calculated as 100\% capacity.

    \item \textbf{Memory Usage (MB):} Total random access memory consumption in megabytes across all participating vehicular nodes during distributed training.

\end{itemize}

\subsection{Baseline Comparisons}

To comprehensively evaluate the proposed adaptive MD-WGAN framework, we compare against state-of-the-art approaches and relevant baseline methods:

\begin{itemize}
    \item \textbf{Random Parameter Selection:} MD-WGAN with random parameter adjustments instead of RL-based optimization, serving as a control for the adaptive decision-making process.
    
    \item \textbf{HVS-GAN \cite{wu2024federated}:} The federated GAN approach proposed in \cite{wu2024federated} which implements federated learning principles for GAN training in IoV environments with centralized parameter aggregation.
    
    \item \textbf{Early Update MD-GAN \cite{Farhoud2025MDGAN}:} Our previously proposed MD-GAN approach with early update strategy and feedback selection mechanism, but without adaptive parameter optimization and game-theoretic coordination. This baseline uses fixed training parameters throughout the simulation and employs asynchronous feedback collection with minimum threshold-based updates.
    
\end{itemize}

These baselines represent different paradigms in distributed GAN training: random adaptation (control), state-of-the-art federated approaches, and communication-efficient methods without adaptive mechanisms. This comparison enables systematic evaluation of the proposed adaptive framework's contributions to both training efficiency and model accuracy in vehicular environments.
We restrict our experimental comparison to GAN-based baselines
because our contribution targets the multi-discriminator,
resource-aware coordination layer within the GAN training process
itself, rather than a comparison of generative modeling paradigms.
Diffusion-based and VAE-based approaches (discussed in
Section~\ref{related}) address the IoV data-synthesis problem with
fundamentally different training dynamics.

\subsection{Reproducibility}
\label{sec:reproducibility}
To support reproducibility, we release our implementation, including the
ns-3 vehicular network simulation, the multi-discriminator
WGAN training pipeline, the reinforcement-learning and game-theoretic
coordination modules, and the configurations used for all baselines and
ablations reported in this paper. The code is publicly available at
\url{https://github.com/farhoudjk/MD-WGAN}.

\subsection{Experimental Results}
\subsubsection{Multi-Objective Performance Evaluation}
\begin{figure}[tbp]
\centerline{\includegraphics[width=0.8\linewidth]{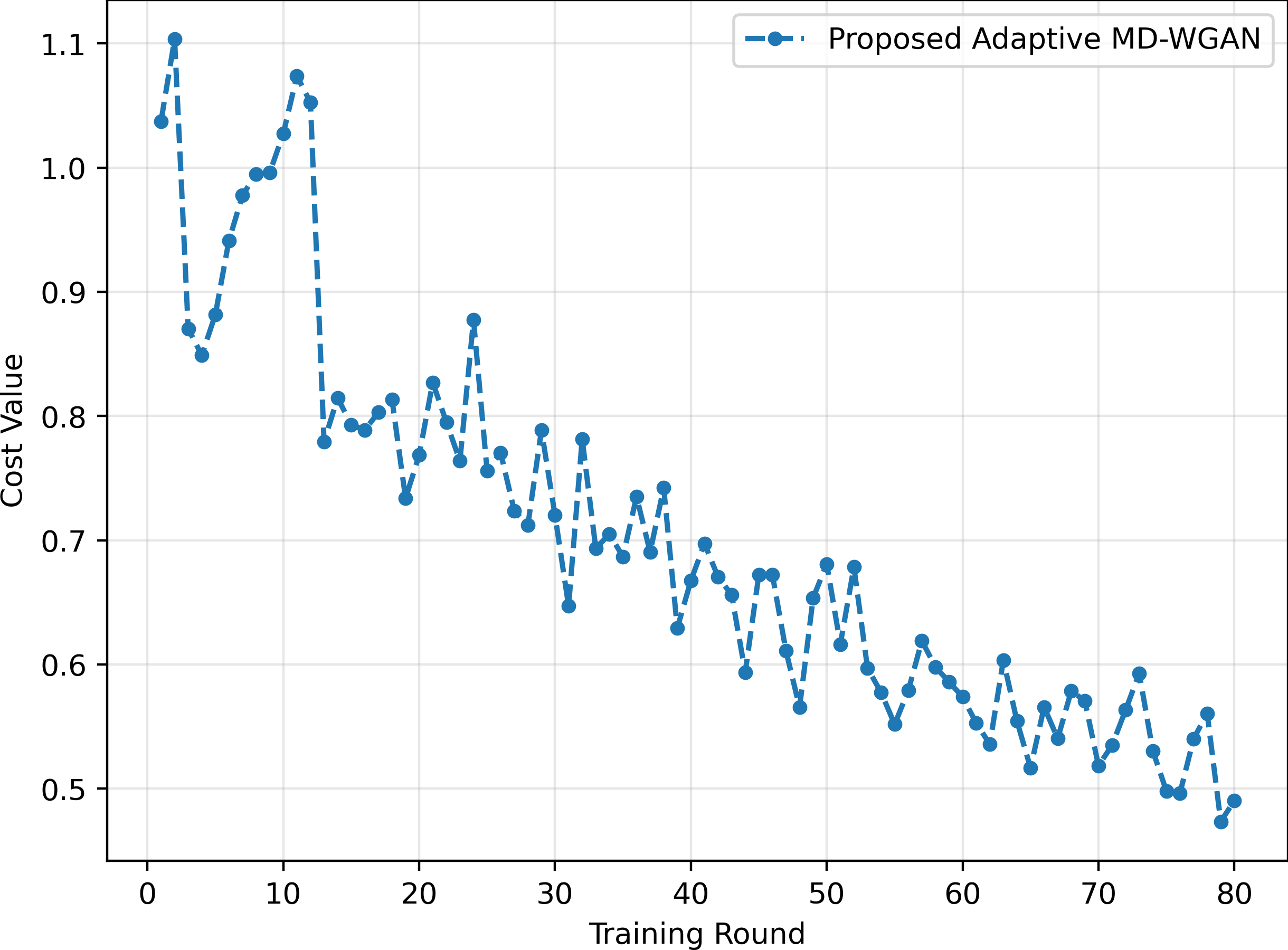}}
\caption{Training \emph{cost} progression of the proposed Adaptive MD-WGAN.} 
\label{fig11}
\end{figure}

To evaluate the overall system performance, we employ a multi-objective cost function that integrates training effectiveness (via sigmoid-transformed GAN loss improvements) with resource efficiency metrics (CPU utilization, memory consumption, communication latency, and network traffic). This comprehensive assessment approach enables quantitative evaluation of competing objectives in the federated IoV environment.
Figure \ref{fig11} illustrates the cost progression across three distinct training phases. The exploration phase (approximately before round 12 or 13) exhibits high volatility with dramatic cost fluctuations between consecutive training iterations. During this initial phase, RL agents employ $\epsilon$-greedy exploration with elevated exploration rates ($\epsilon_{decay}$ = 0.995), resulting in random discriminator selections that produce both substantial performance gains and significant degradations. The sigmoid term dominates the cost calculation due to large GAN loss variations, while resource utilization remains suboptimal.
The learning phase (approximately rounds 13-57) demonstrates the framework's adaptive capabilities as RL agents progressively acquire optimal policies for discriminator selection and resource coordination. Cost stability improves markedly while maintaining consistent downward trends, indicating successful policy convergence and effective best-response coordination. 
The optimization phase (approximately after round 57) achieves stable, near-optimal performance with minimal round-to-round fluctuations. As the system approaches convergence, the sigmoid transformation approaches zero due to negligible GAN loss improvements, effectively shifting optimization focus toward resource efficiency maximization.

\subsubsection{Accuracy Performance Analysis}
\begin{figure}[tbp]
\centerline{\includegraphics[width=1\linewidth]{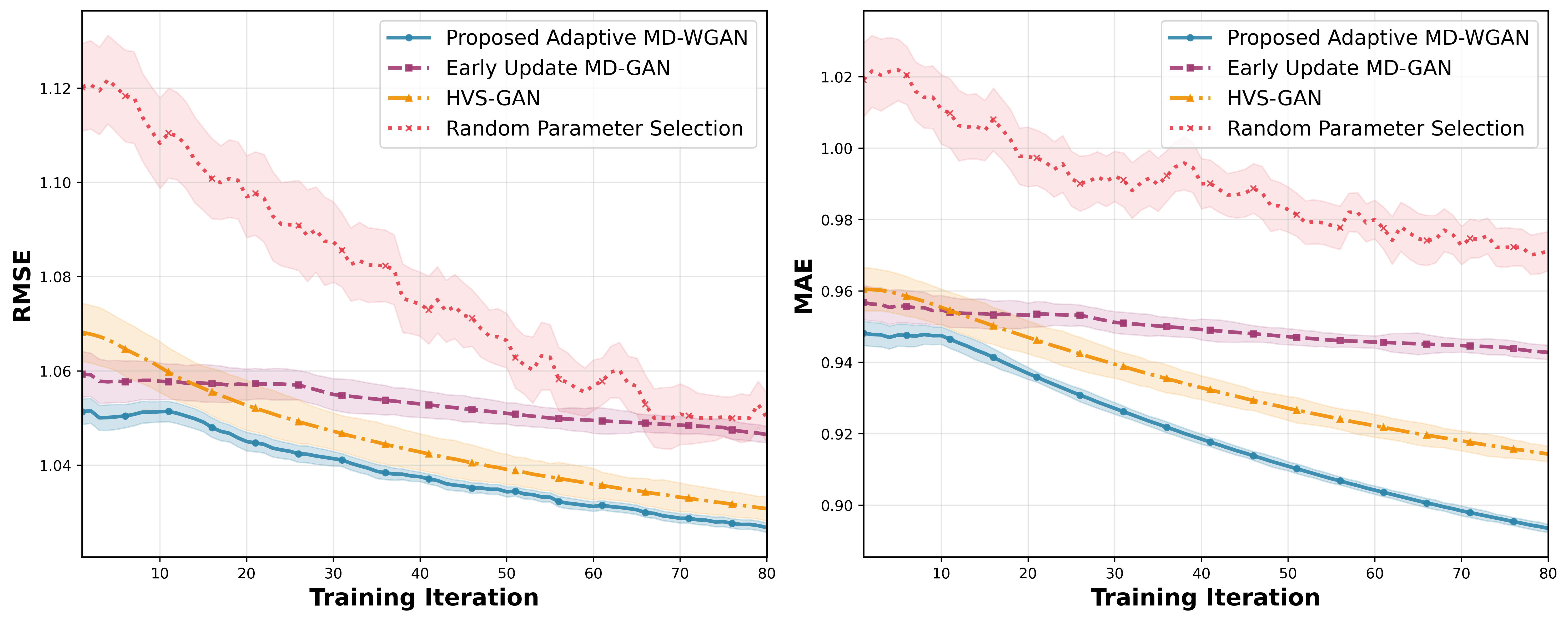}}
\caption{Performance comparison of the proposed approach in terms of RMSE and MAE by increasing the training iteration} 
\label{fig2}
\end{figure}

Figure~\ref{fig2} demonstrates the convergence characteristics of all evaluated methods over 80 training iterations. The proposed adaptive MD-WGAN framework demonstrates better convergence performance, achieving the lowest RMSE (1.029) and MAE (0.894) values while maintaining stable training throughout 80 iterations. This better performance stems from the dual-zone architecture that enables real-time parameter optimization through reinforcement learning agents and game-theoretic resource allocation, allowing the system to continuously adapt learning rates, batch sizes, and communication intervals based on current network conditions. Unlike fixed-parameter approaches that plateau or show delayed convergence, the adaptive framework maintains consistent improvement throughout the training process, demonstrating the critical importance of intelligent parameter adaptation in dynamic vehicular environments.

In contrast, the baseline methods exhibit significant limitations that highlight the advantages of the proposed approach. HVS-GAN, while showing steady improvement to RMSE = 1.032, suffers from centralized parameter aggregation that introduces communication overhead and fails to adapt to local network conditions. The Early Update MD-GAN approach, despite its communication efficiency benefits, demonstrates initial plateau behavior before gradual improvement (final RMSE = 1.044), indicating that static parameters cannot optimally respond to varying network conditions and data characteristics. The random parameter selection baseline validates the necessity of intelligent adaptation, showing the worst performance (RMSE = 1.078) with high instability and erratic convergence patterns. The better performance of the adaptive framework results from its ability to prevent training stagnation through continuous parameter optimization, eliminate resource conflicts via best-response coordination, and enable localized optimization while maintaining global coherence, making it particularly suited for the dynamic and resource-constrained nature of vehicular networks.

\subsubsection{Resource Utilization}
\begin{figure}[tbp]
\centerline{\includegraphics[width=0.98\linewidth]{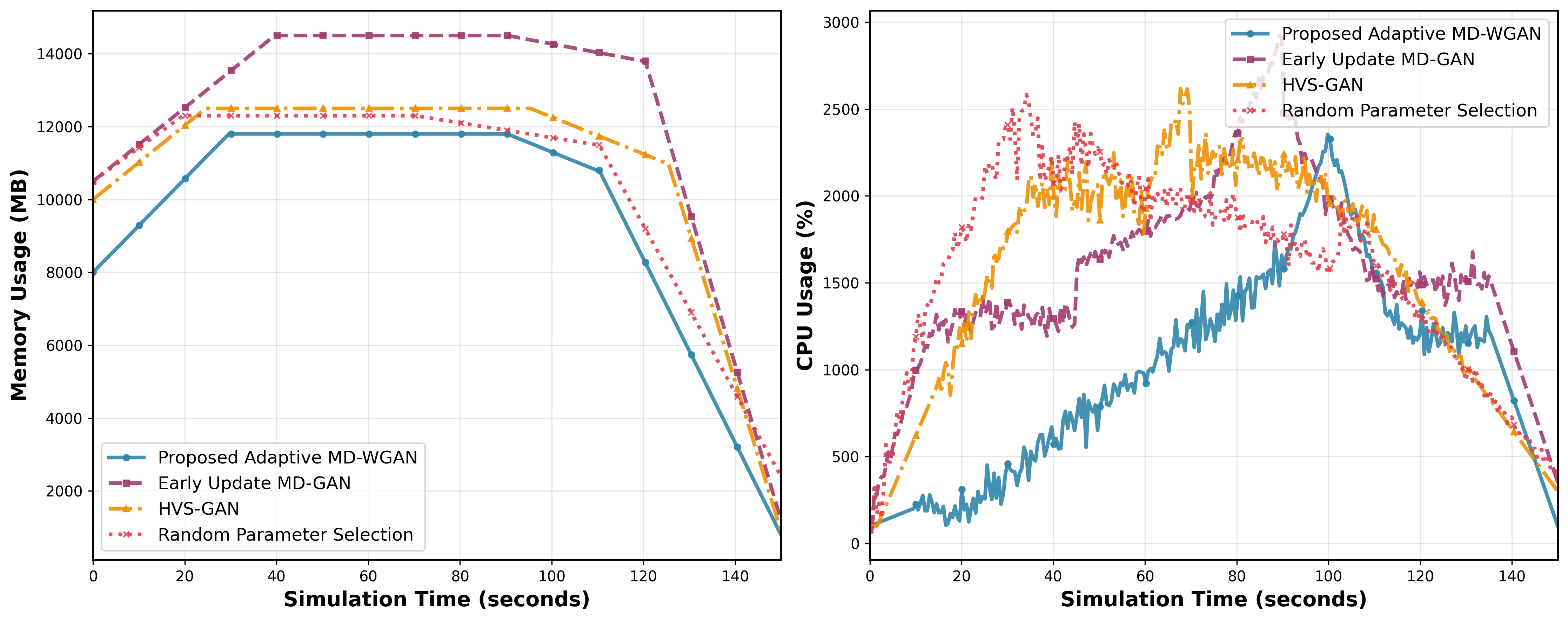}}
\caption{ Performance Comparison of the proposed approach in terms of CPU and memory usage during simulation time} 
\label{fig3}
\end{figure}

Figure~\ref{fig3} presents the memory and CPU usage patterns of all evaluated methods throughout the 150-second simulation period. 
The proposed adaptive MD-WGAN exhibits the most efficient memory utilization pattern, maintaining stable usage around 8,000-12,000 MB during the active training phase (0-100 seconds) before efficiently releasing resources as training concludes. In contrast, the Early Update MD-GAN shows consistently high memory consumption, peaking at approximately 14,500 MB and maintaining elevated usage throughout the simulation. The reason for this is that all the discriminators are working simultaneously in the networks, requiring substantial memory allocation to maintain multiple discriminator instances and their associated computational graphs in memory concurrently. HVS-GAN demonstrates moderate memory efficiency with usage ranging between 10,000-12,500 MB, while the Random Parameter Selection approach shows erratic memory allocation patterns with high variance, reflecting the inefficiency of arbitrary parameter adjustments. 

The CPU utilization patterns reveal significant differences in computational efficiency across methods. The proposed adaptive framework maintains stable CPU usage between 500-1,400\% during training, demonstrating efficient parallel processing and resource coordination through the dual-zone architecture. The method shows controlled fluctuations that correspond to adaptive parameter adjustments, indicating responsive resource allocation based on training requirements. Early Update MD-GAN and HVS-GAN exhibit higher and more erratic CPU usage patterns, with frequent spikes reaching 1,500-2,200\%, suggesting inefficient resource allocation and potential processing bottlenecks. The Random Parameter Selection baseline displays the most chaotic CPU behavior with extreme variability and frequent spikes above 1,500\%, confirming the computational overhead of uncoordinated parameter changes. The proposed method's ability to maintain lower average CPU usage while achieving superior accuracy demonstrates the effectiveness of the game-theoretic resource allocation mechanism in preventing computational conflicts and optimizing resource distribution among competing training processes. Additionally, the CPU usage peaks exceeding 1600\% can be attributed to hyperthreading capabilities in EC2 instances\cite{fenn2010performance}, which allow each physical core to handle multiple threads simultaneously, leading to reported CPU utilization values that surpass the theoretical maximum based on the number of virtual CPUs.

\subsubsection{Network Traffic Overhead}
\begin{figure}[tbp]
\centerline{\includegraphics[width=0.785\linewidth]{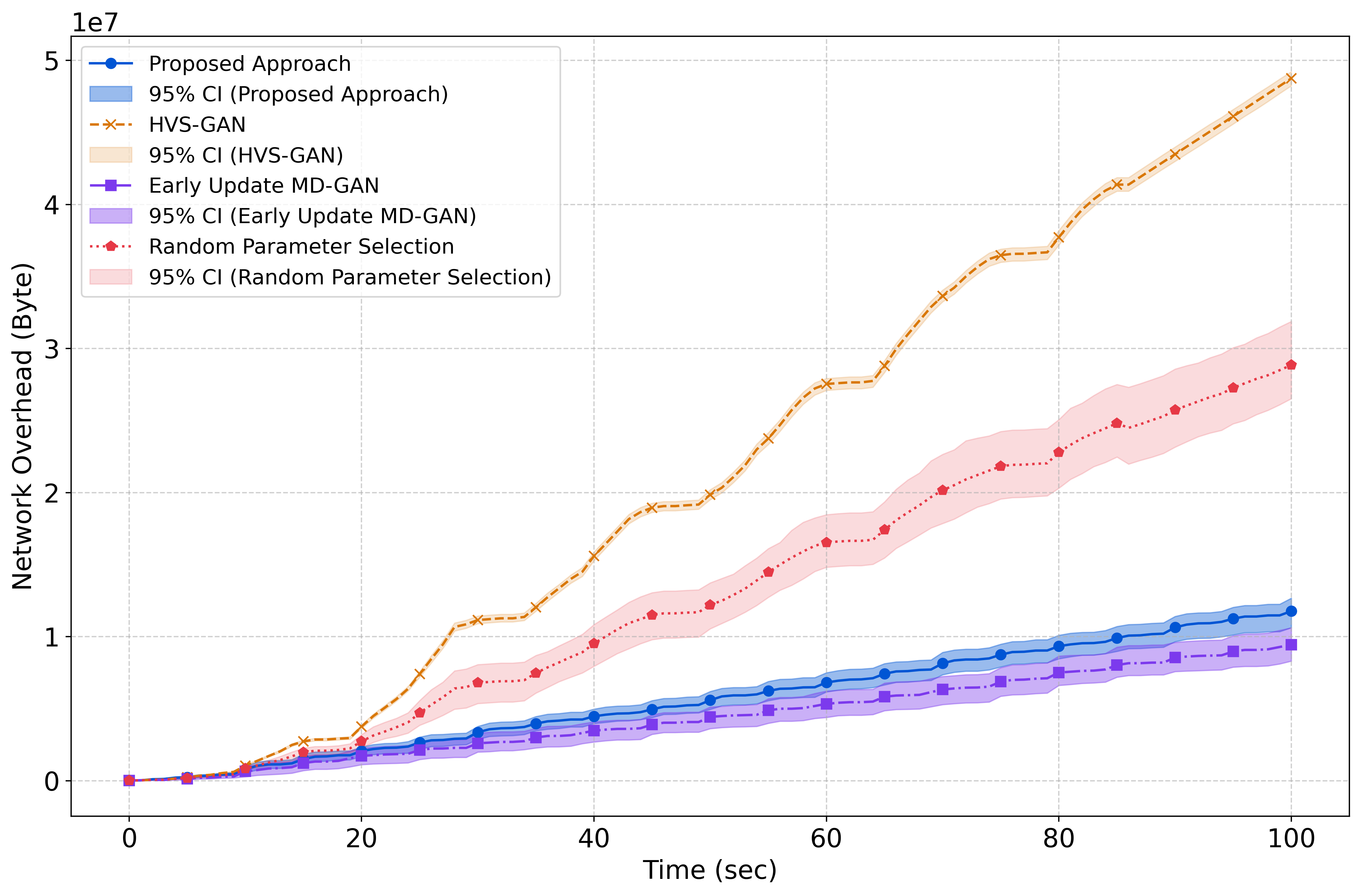}}
\caption{Comparison of the proposed method in terms of Network Overhead} 
\label{fig4}
\end{figure}

Figure~\ref{fig4} shows the comparative network overhead performance of the different GAN-based approaches over a 100-second simulation period. The results demonstrate a clear performance hierarchy, with the Early Update MD-GAN method achieving the best performance by maintaining the lowest network overhead throughout the simulation, reaching only approximately 0.9 million bytes by the end of the 100-second period. This superior performance is primarily attributed to its feedback selection mechanism, which prevents nodes exceeding predefined thresholds from transmitting updates, thereby significantly reducing unnecessary network traffic. The proposed approach follows as the second-best performer, maintaining relatively low and controlled network overhead of approximately 1.2 million bytes with steady, linear growth characteristics. While the proposed method exhibits slightly higher overhead than the Early Update MD-GAN, this marginal increase should be considered within the context of resource usage tradeoffs, where the additional network cost may be offset by improved computational efficiency and model performance gains. In contrast, the HVS-GAN approach demonstrates the poorest performance with dramatic exponential growth culminating in nearly 4.8 million bytes of network overhead, which can be attributed to its reliance on a centralized controller for reinforcement learning operations that must coordinate with selected nodes and collect results from each participating node, creating substantial communication bottlenecks. The Random Parameter Selection method also shows problematic exponential growth reaching approximately 3.1 million bytes, highlighting how the absence of intelligent coordination mechanisms leads to inefficient resource utilization and excessive network overhead in vehicular network environments.

\subsubsection{Latency}

\begin{figure}[tbp]
\centerline{\includegraphics[width=0.75\linewidth]{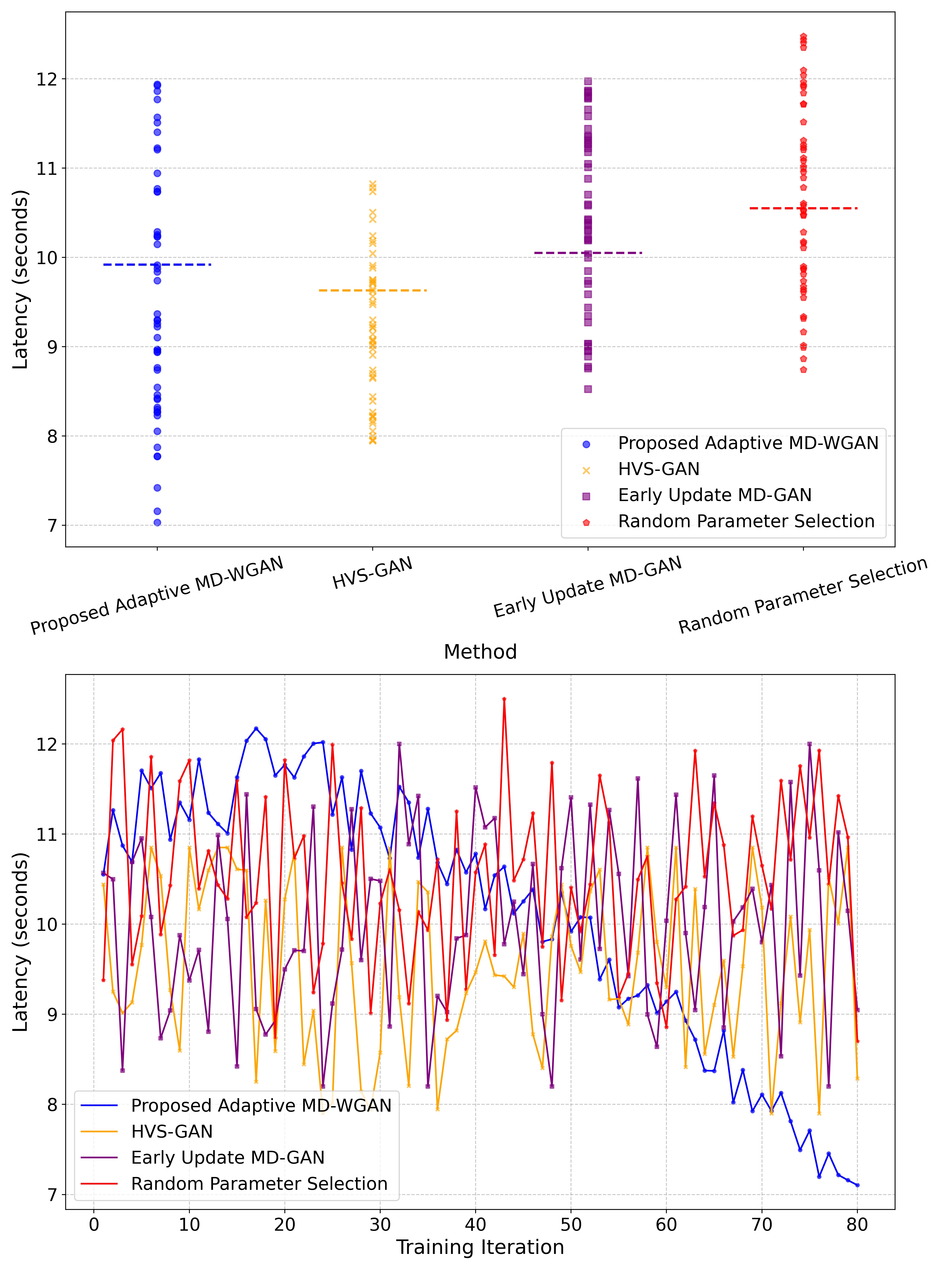}}
\caption{Performance Comparison of the proposed approach in terms of Training Latency} 
\label{fig5}
\end{figure}

Figure~\ref{fig5} shows the latency performance comparison of the four GAN-based approaches for IoV applications, presented through both box plot distribution analysis (top) and training iteration progression (bottom). The box plot reveals that the proposed approach achieves a stable performance, with a median latency of approximately 9.9 seconds and the tightest distribution range, indicating a good stability and predictability in vehicular network conditions. HVS-GAN demonstrates best performance with a median around 9.7 seconds but exhibits greater variability over training round, while Early Update MD-GAN shows a median of approximately 10.1 seconds with considerable spread in latency values. Random Parameter Selection displays the poorest consistency with the highest median latency of about 10.6 seconds and the widest distribution range, reflecting the inefficiency of non-adaptive parameter management. The training iteration progression chart further illustrates these performance characteristics, showing that while all methods begin with similar latency values around 8-12 seconds, the proposed approach demonstrates a distinctive downward trend in latency as training progresses, particularly evident after iteration 50 where it achieves significantly lower latency values approaching 7 seconds. This superior latency reduction capability can be attributed to the inclusion of latency as a specific parameter in the Q-learning cost function, enabling the system to actively optimize for reduced response times, whereas the other approaches exhibit normal fluctuating behavior without systematic latency improvement throughout the training process.
It shows that as training progresses over time, the latency of our proposed work consistently decreases. This is a direct consequence of our Q-learning cost function, which includes a specific parameter designed to penalize high latency. In contrast, the latency for all other methods fluctuates randomly without any clear optimization trend, indicating their inability to actively adapt to and mitigate network delays. This provides compelling evidence that our framework is not just passively low-latency but possesses an intelligent, active optimization mechanism that is essential for real-world, dynamic IoV environments.

\subsubsection{Scalability Analysis}

The scalability analysis is based on the following variables:
\begin{itemize}
    \item $N_G$: The total number of generator nodes (vehicles).
    \item $K$: The total number of discriminators.
    \item $|\mathcal{S}|$: The dimensionality of the state space for the RL agent.
    \item $|\mathcal{A}|$: The dimensionality of the action space for the RL agent.
    \item $E_{epochs}$: The total number of training epochs.
    \item $d$: The total number of parameters in the generator or a single discriminator.
    \item $B$: The batch size used for training.
\end{itemize}

\textbf{Adaptive Configuration Complexity:}

The \textit{Adaptive Configuration} phase involves multi-agent RL and game-theoretic optimization. The computational cost of these components determines the overhead of the adaptive decision-making process.

\begin{itemize}
 \item \textit{Multi-Agent Reinforcement Learning (DQN):}
The complexity of a single training step for the DQN agent is dominated by the neural network's forward and backward passes. Given the network structure of $[128, 128, |\mathcal{A}|]$, where 128 represents the number of neurons in each hidden layer, the complexity is:
\[
T_{\text{RL}} = \mathcal{O}(|\mathcal{S}| \cdot 128 + 128 \cdot 128 + 128 \cdot |\mathcal{A}|)
\]
Since each generator's local discriminator pool size $K_{\text{local}}$
is bounded by design (Table~\ref{tab:sim_params}) independent of the
total number of discriminators $K$ in the network -- scaling the
system adds RSUs rather than growing any single RSU's local pool --
the per-decision-cycle DQN complexity remains bounded as the network
scales, i.e., $T_{\mathrm{RL}} = \mathcal{O}(K_{\text{local}}) = \mathcal{O}(1)$
with respect to $N_G$ and $K$.

\item \textit{Game-Theoretic Training Optimization:} The coordination step evaluates
best responses over the fixed discrete strategy sets $\mathcal{E}_D$ and
$\mathcal{E}_G$ in two passes. Its cost is therefore
$\mathcal{O}(|\mathcal{E}_D|\cdot|\mathcal{E}_G|)$. Because both sets are bounded
by design (Table~\ref{tab:nash_hyperparams}), this is constant with respect to
network scale:
\begin{equation}
    T_{\text{coord}} = \mathcal{O}\big(|\mathcal{E}_D|\cdot|\mathcal{E}_G|\big)
    = \mathcal{O}(1).
\end{equation}
\end{itemize}
\textbf{Distributed Training and Synchronization Complexity}

This is the most critical phase for scalability analysis, as its complexity directly depends on the number of nodes, discriminators, and data size.

\subsubsection*{Computational Complexity ($T_{\text{train}}$)}
The time complexity per training epoch is determined by the most computationally intensive task, which is the gradient calculation and backpropagation for the generator and discriminators.

\begin{itemize}
    \item Discriminator Optimization ($\mathcal{L}_{D_j}$): Each of the $K$ discriminators processes a batch of size $B$. The time complexity to compute the loss and gradients scales with the number of parameters $d$. Since these operations occur in parallel across different nodes, the time for this step is a function of the model size and batch size, not the number of nodes.
    \[
    T_{D_{\text{train}}} = \mathcal{O}(d \cdot B)
    \]
    \item Generator Optimization ($\mathcal{L}_G$): The generator receives feedback from the ensemble of discriminators. Its loss calculation and gradient update also scale with the number of parameters and the batch size.
    \[
    T_{G_{\text{train}}} = \mathcal{O}(d \cdot B)
    \]
\end{itemize}
The overall computational complexity per training epoch, assuming parallel execution, is:
\[
T_{\text{train}} = \mathcal{O}(d \cdot B)
\]
The total computational complexity is the number of training epochs multiplied by the complexity per epoch:
\[
T_{\text{total}} = E_{\text{epochs}} \cdot T_{\text{train}} = \mathcal{O}(E_{\text{epochs}} \cdot d \cdot B)
\]

\subsubsection*{Communication Complexity ($C_{\text{comm}}$)}
Communication is a major bottleneck. The complexity is determined by the amount of data exchanged for parameter aggregation.

\begin{itemize}
    \item Generator Sync: The generator parameters are sent from each of the $N_G$ generator nodes. The total data transferred scales with the number of generator nodes and the number of parameters.
    \[
    C_{G_{\text{sync}}} = \mathcal{O}(N_G \cdot d)
    \]
    \item Discriminator Sync:  Each of the $K_S$ discriminators ($K_S \le K$) sends its gradients, with a size of $d$, back to the generator. The cost for a single generator to receive and aggregate these gradients is:
$$\text{Cost}_{\text{Agg}} = \mathcal{O}(K_S \cdot d)$$

The total network cost for this phase is the sum of the communication costs for all $N_G$ generator nodes operating in parallel. By combining the costs of sending parameters and receiving gradients, the overall complexity becomes:
$$C_{\text{Dsync}} = \mathcal{O}(N_G \cdot K_S \cdot d)$$
\end{itemize}
The total communication complexity per synchronization step is the sum of these components:
\[
C_{\text{comm}} = C_{G_{\text{sync}}} + C_{D_{\text{sync}}} = \mathcal{O}(N_G \cdot K_S \cdot d)
\]
The total communication traffic is the number of synchronization steps multiplied by the complexity per step:
\[
C_{\text{total}} = S_{\text{sync}} \cdot C_{\text{comm}} = \mathcal{O}(S_{\text{sync}} \cdot N_G \cdot K_S \cdot d)
\]


Table \ref{tab:scalability_summary} summarizes the scalability of the proposed framework in terms of its key resource dimensions.

\begin{table}[H]
    \centering
    \caption{Summary of Scalability Analysis}
    \label{tab:scalability_summary}
    \begin{tabular}{p{1.6cm}p{1.8cm}p{4.3cm}}
        \toprule
        \textbf{Aspect} & \textbf{Complexity} & \textbf{Implications} \\
        \midrule
        \textbf{Computational} & $\mathcal{O}(E_{\text{epochs}} \cdot d \cdot B)$ & Scales with training parameters but is "parallelizable" across nodes. Adding more nodes does not increase this complexity. \\
        \midrule
        \textbf{Communication} & $\mathcal{O}(S_{\text{sync}} \cdot N_G \cdot K_S \cdot d)$ &  The communication overhead scales linearly with the number of generator nodes ($N_G$) and its related discriminators ($K_S$). The framework's success hinges on adaptive strategies (such as Intelligent Communication Scheduling) to minimize the number of synchronization steps ($S_{\text{sync}}$) and the number of active discriminators ($|K|$) per generator. \\
        \bottomrule
    \end{tabular}
\end{table}

\begin{figure}[tbp]
\centerline{\includegraphics[width=1\linewidth]{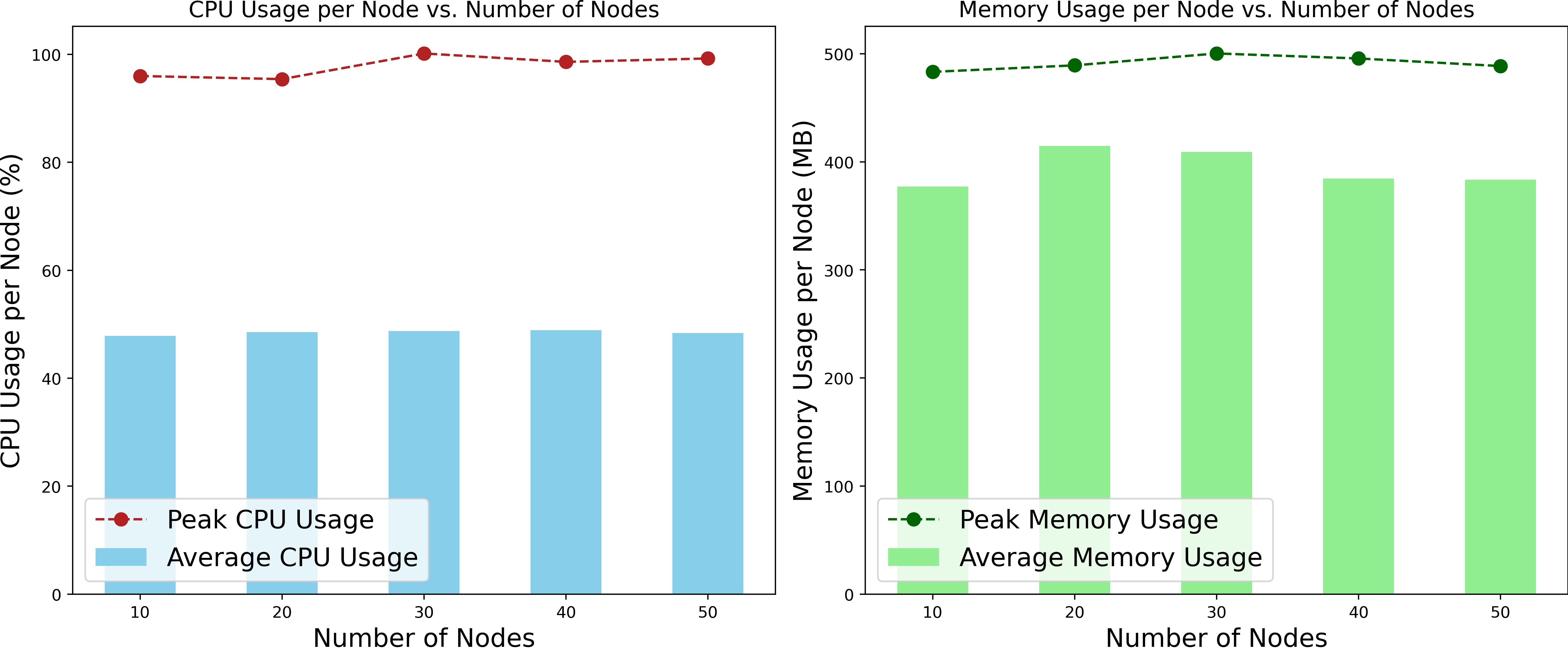}}
\caption{Computational scalability of the proposed approach} 
\label{fig6}
\end{figure}

\begin{figure}[tbp]
\centerline{\includegraphics[width=0.72\linewidth]{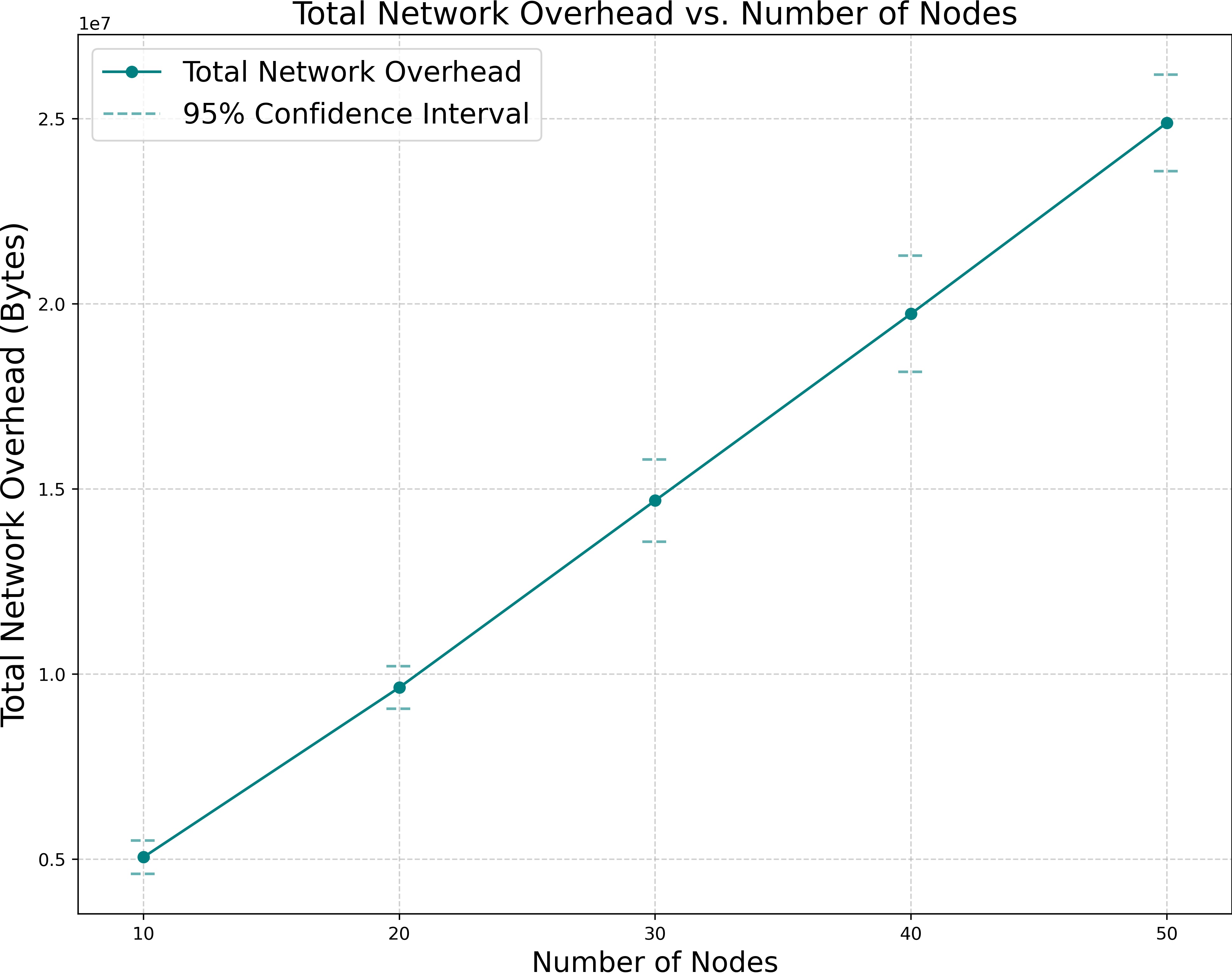}}
\caption{Communication scalability of the proposed approach} 
\label{fig7}
\end{figure}

To support our scalability analysis, we also calculated the network overhead and resource usage as we increased the number of nodes in the IoV environment (Figures~\ref{fig6} and~\ref{fig7}). This experimental validation demonstrates how our proposed framework manages computational and communication costs as the system scales.

\subsubsection{Application-Specific Multi-Objective Performance Validation}
\label{sec:app-specific}
To validate the effectiveness of our proposed multi-objective cost function in adapting to diverse IoV applications, we define four distinct task scenarios. Each scenario prioritizes a different performance metric by adjusting the weighting coefficients ($\alpha_G, \alpha_1, \alpha_2, \alpha_3, \alpha_4$) in the cost function, ensuring their sum remains constant (equal to 1). This controlled experimental design allows us to isolate and measure the impact of each objective on the overall framework performance.
\begin{figure*}[tbp]
\centerline{\includegraphics[width=0.7\linewidth]{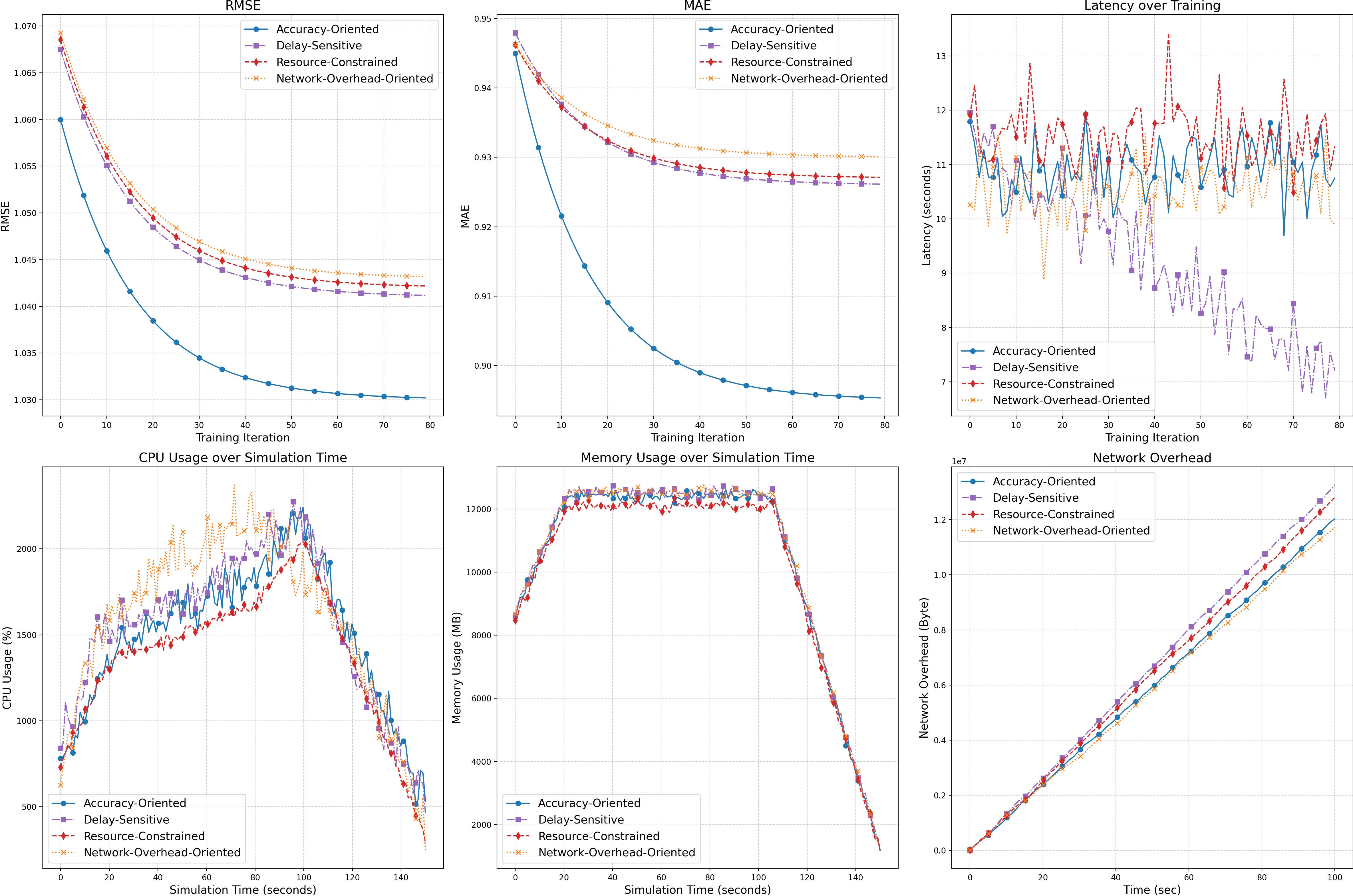}}
\caption{Comparative performance of the proposed approach for different apps in terms of varying metrics} 
\label{fig8}
\end{figure*}

For each scenario, we configure the reinforcement learning agent's cost function to prioritize a specific performance goal. The coefficient values for the four scenarios are defined as follows:

\begin{enumerate}
    \item \textbf{Delay-Sensitive Application}: This scenario prioritizes minimizing communication latency, which is crucial for safety-critical IoV tasks like autonomous braking. We set the latency coefficient $\boldsymbol{\alpha_3}$ to a high value of \textbf{0.4}, while the other coefficients ($\alpha_G, \alpha_1, \alpha_2, \alpha_4$) are each set to \textbf{0.15}. This configuration heavily rewards the agent for actions that reduce network delay.

    \item \textbf{Accuracy-Oriented Application}: Here, the primary objective is to maximize the GAN's generation quality, which is vital for high-fidelity data synthesis in applications such as sensor data fusion or environmental modeling. The GAN objective coefficient $\boldsymbol{\alpha_G}$ is set to \textbf{0.4}, with the remaining coefficients ($\alpha_1, \alpha_2, \alpha_3, \alpha_4$) set to \textbf{0.15} each. This configuration incentivizes the agent to prioritize training stability and convergence.

    \item \textbf{Resource-Constrained Application}: This scenario focuses on minimizing the consumption of limited computational resources, essential for IoV nodes with low-power CPUs and limited memory. The coefficients for CPU ($\boldsymbol{\alpha_1}$) and memory ($\boldsymbol{\alpha_2}$) are each set to \textbf{0.3}, while the remaining coefficients are set to \textbf{0.133} each that pushes the agent to adopt resource-efficient strategies.

    \item \textbf{Network Overhead-Oriented Application}: The goal here is to minimize the total network traffic, which is critical in environments with limited bandwidth. The network traffic coefficient $\boldsymbol{\alpha_4}$ is set to \textbf{0.4}, with the other coefficients ($\alpha_G, \alpha_1, \alpha_2, \alpha_3$) at \textbf{0.15} each. This configuration encourages the agent to prioritize data compression and intelligent communication scheduling.
\end{enumerate}

Figure~\ref{fig8} presents the comparative performance of our proposed framework under four distinct application scenarios. Each subplot corresponds to a key performance metric.
The results validate the cost function's ability to adapt the framework's behavior to specific task requirements.

\section{Conclusion} \label{conclusion}
This paper presents a novel adaptive multi-discriminator Wasserstein GAN framework that successfully integrates reinforcement learning with game-theoretic coordination to achieve simultaneous optimization of training accuracy, resource efficiency, communication overhead, and latency constraints in IoV networks.
The key contributions include a unified optimization framework consolidating adversarial learning with practical deployment constraints, a hierarchical coordination protocol combining Deep Q-Network agents with game-theoretic best-response coordination, and a dual-zone architecture enabling real-time parameter adaptation. Experimental validation demonstrates better performance with improvement in prediction accuracy, reduction in resource consumption and consistent latency optimization compared to state-of-the-art approaches.
The demonstrated effectiveness establishes a foundation for next-generation AI-driven vehicular network optimization, contributing to intelligent transportation systems advancement through improved efficiency, reliability, and real-time performance. The framework provides a practical solution for vehicular network operators and ITS developers seeking to leverage machine learning capabilities while maintaining operational efficiency, making it particularly valuable for connected and autonomous vehicles, smart city infrastructure, and cooperative transportation systems.

Future research directions include investigating advanced compression techniques, extending support for emerging 5G/6G standards, and developing specialized adaptations for autonomous vehicle coordination and smart traffic management.



\ifCLASSOPTIONcaptionsoff
  \newpage
\fi



%
\bibliographystyle{IEEEtran}
\bibliography{bibtex/IEEEabrv,bibtex/IEEEexample}

\begin{thebibliography}{10}
\providecommand{\url}[1]{#1}
\csname url@samestyle\endcsname
\providecommand{\newblock}{\relax}
\providecommand{\bibinfo}[2]{#2}
\providecommand{\BIBentrySTDinterwordspacing}{\spaceskip=0pt\relax}
\providecommand{\BIBentryALTinterwordstretchfactor}{4}
\providecommand{\BIBentryALTinterwordspacing}{\spaceskip=\fontdimen2\font plus
\BIBentryALTinterwordstretchfactor\fontdimen3\font minus \fontdimen4\font\relax}
\providecommand{\BIBforeignlanguage}[2]{{%
\expandafter\ifx\csname l@#1\endcsname\relax
\typeout{** WARNING: IEEEtran.bst: No hyphenation pattern has been}%
\typeout{** loaded for the language `#1'. Using the pattern for}%
\typeout{** the default language instead.}%
\else
\language=\csname l@#1\endcsname
\fi
#2}}
\providecommand{\BIBdecl}{\relax}
\BIBdecl

\bibitem{pratap2023cotev}
S.~Pratap, P.~Dass, and S.~Misra, ``Cotev: trustworthy and cooperative task execution in internet of vehicles,'' \emph{IEEE Transactions on Mobile Computing}, vol.~23, no.~4, pp. 2915--2926, 2023.

\bibitem{devika2024vadgan}
S.~Devika, R.~R. Shrivastava, P.~Narang, T.~Alladi, and F.~R. Yu, ``Vadgan: An unsupervised gan framework for enhanced anomaly detection in connected and autonomous vehicles,'' \emph{IEEE Transactions on Vehicular Technology}, vol.~73, no.~9, pp. 12\,458--12\,467, 2024.

\bibitem{zhang2023novel}
J.~Zhang, L.~Zhao, K.~Yu, G.~Min, A.~Y. Al-Dubai, and A.~Y. Zomaya, ``A novel federated learning scheme for generative adversarial networks,'' \emph{IEEE Transactions on Mobile Computing}, vol.~23, no.~5, pp. 3633--3649, 2023.

\bibitem{mishra2023collaborative}
K.~Mishra, G.~N. Rajareddy, U.~Ghugar, G.~S. Chhabra, and A.~H. Gandomi, ``A collaborative computation and offloading for compute-intensive and latency-sensitive dependency-aware tasks in dew-enabled vehicular fog computing: A federated deep q-learning approach,'' \emph{IEEE Transactions on Network and Service Management}, vol.~20, no.~4, pp. 4600--4614, 2023.

\bibitem{farhoudtsc}
F.~J. Kaleibar, M.~St-Hilaire, and M.~Barati, ``A novel architectural framework for semantic-based service provisioning in vehicular cloud networks,'' \emph{IEEE Transactions on Services Computing}, pp. 1--14, 2026.

\bibitem{mfogo2025gan}
V.~S. Mfogo, A.~Zemkoho, L.~Njilla, M.~J. Nkenlifack, and C.~A. Kamhoua, ``Gan-aiipot: Gan-based cyber deception for probing attacks on iot devices,'' \emph{IEEE Transactions on Network and Service Management}, vol.~23, pp. 417--431, 2025.

\bibitem{luo2025high}
X.~Luo, Z.~Wei, G.~Zhang, H.~Huang, and R.~Zhou, ``High-risk powered two-wheelers scenarios generation for autonomous vehicle testing using wgan,'' \emph{Traffic Injury Prevention}, vol.~26, no.~2, pp. 243--251, 2025.

\bibitem{amrvec}
A.~M. Zaki, S.~A. Elsayed, K.~Elgazzar, H.~S. Hassanein, and S.~Moussa, ``Mambavec: A multi-agent drl framework with mamba for efficient long-term vehicular edge computing task offloading,'' in \emph{2026 International Conference on Smart Mobility (SM)}, 2026, pp. 1--7.

\bibitem{contreras2025generative}
J.~Contreras-Castillo, J.~Guerrero-Iba{\~n}ez, S.~Zeadally, and E.-K. Hong, ``Generative ai for internet of vehicles (iov): Potential and challenges,'' \emph{IEEE Communications Standards Magazine}, 2025.

\bibitem{jafari2021sla}
F.~J. Kaleibar and M.~St-Hilaire, ``An adaptive clustering approach for dynamic service provisioning in vehicular cloud networks,'' in \emph{2024 IEEE 10th World Forum on Internet of Things (WF-IoT)}.\hskip 1em plus 0.5em minus 0.4em\relax IEEE, 2024, pp. 666--671.

\bibitem{hammoud2022dynamic}
A.~Hammoud, M.~Kantardjian, A.~Najjar, A.~Mourad, H.~Otrok, Z.~Dziong, and N.~Guizani, ``Dynamic fog federation scheme for internet of vehicles,'' \emph{IEEE Transactions on Network and Service Management}, vol.~20, no.~2, pp. 1913--1923, 2022.

\bibitem{yuan2020dynamic}
T.~Yuan, W.~da~Rocha~Neto, C.~E. Rothenberg, K.~Obraczka, C.~Barakat, and T.~Turletti, ``Dynamic controller assignment in software defined internet of vehicles through multi-agent deep reinforcement learning,'' \emph{IEEE Transactions on Network and Service Management}, vol.~18, no.~1, pp. 585--596, 2020.

\bibitem{karim2022architecture}
S.~M. Karim, A.~Habbal, S.~A. Chaudhry, and A.~Irshad, ``Architecture, protocols, and security in iov: Taxonomy, analysis, challenges, and solutions,'' \emph{Security and Communication Networks}, vol. 2022, no.~1, p. 1131479, 2022.

\bibitem{elbir2022federated}
A.~M. Elbir, B.~Soner, S.~{\c{C}}{\"o}leri, D.~G{\"u}nd{\"u}z, and M.~Bennis, ``Federated learning in vehicular networks,'' in \emph{2022 IEEE International Mediterranean Conference on Communications and Networking (MeditCom)}.\hskip 1em plus 0.5em minus 0.4em\relax IEEE, 2022, pp. 72--77.

\bibitem{wang2022joint}
G.~Wang, F.~Xu, H.~Zhang, and C.~Zhao, ``Joint resource management for mobility supported federated learning in internet of vehicles,'' \emph{Future Generation Computer Systems}, vol. 129, pp. 199--211, 2022.

\bibitem{posner2021federated}
J.~Posner, L.~Tseng, M.~Aloqaily, and Y.~Jararweh, ``Federated learning in vehicular networks: Opportunities and solutions,'' \emph{IEEE Network}, vol.~35, no.~2, pp. 152--159, 2021.

\bibitem{kaleibar2025customized}
F.~J. Kaleibar, M.~St-Hilaire, and M.~Barati, ``A customized genetic algorithm for sla-aware service provisioning in infrastructure-less vehicular cloud networks,'' \emph{IEEE Transactions on Services Computing}, vol.~18, pp. 499--512, 2025.

\bibitem{wang2022federated}
W.~Wang, C.~Liang, L.~Tang, H.~Yanikomeroglu, and Q.~Chen, ``Federated multi-discriminator biwgan-gp based collaborative anomaly detection for virtualized network slicing,'' \emph{IEEE Transactions on Mobile Computing}, vol.~22, no.~11, pp. 6445--6459, 2022.

\bibitem{wu2024federated}
L.~Wu, H.~Lin, and X.~Wang, ``Federated training generative adversarial networks for heterogeneous vehicle scheduling in iov,'' \emph{IEEE Internet of Things Journal}, 2024.

\bibitem{xie2024gai}
G.~Xie, Z.~Xiong, X.~Zhang, R.~Xie, S.~Guo, M.~Guizani, and H.~V. Poor, ``Gai-iov: Bridging generative ai and vehicular networks for ubiquitous edge intelligence,'' \emph{IEEE Transactions on Wireless Communications}, vol.~23, no.~10, pp. 12\,799--12\,814, 2024.

\bibitem{quan2024federated}
H.~Quan, Q.~Zhang, and J.~Zhao, ``Federated learning assisted intelligent iov mobile edge computing,'' \emph{IEEE Transactions on Green Communications and Networking}, vol.~9, no.~1, pp. 228--241, 2024.

\bibitem{hazarika2023multi}
B.~Hazarika, K.~Singh, S.~Biswas, S.~Mumtaz, and C.-P. Li, ``Multi-agent drl-based task offloading in multiple ris-aided iov networks,'' \emph{IEEE Transactions on Vehicular Technology}, vol.~73, no.~1, pp. 1175--1190, 2023.

\bibitem{xu2024multi}
X.~Xu, L.~Gu, M.~Bilal, M.~Khan, Y.~Wen, G.~Liu, and Y.~Yuan, ``Multi-agent reinforcement learning based edge content caching for connected autonomous vehicles in iov,'' \emph{ACM Transactions on Autonomous and Adaptive Systems}, 2024.

\bibitem{Farhoud2025MDGAN}
F.~Jafari~Kaleibar and M.~Litoiu, ``A distributed gan-based framework for low-latency and efficient learning in the internet of vehicles,'' in \emph{33rd International Symposium on the Modeling, Analysis, and Simulation of Computer and Telecommunication System (MASCOTS2025)}, 2025, pp. 1--8.

\bibitem{liang2020reinforcement}
H.~Liang, X.~Zhang, X.~Hong, Z.~Zhang, M.~Li, G.~Hu, and F.~Hou, ``Reinforcement learning enabled dynamic resource allocation in the internet of vehicles,'' \emph{IEEE Transactions on Industrial Informatics}, vol.~17, no.~7, pp. 4957--4967, 2020.

\bibitem{zaki2025cogroup}
A.~M. Zaki, S.~A. Elsayed, K.~Elgazzar, and H.~S. Hassanein, ``Cogroup: Cooperative quality offloading with worker grouping using hierarchical multi-agent deep reinforcement learning,'' in \emph{2025 International Wireless Communications and Mobile Computing (IWCMC)}.\hskip 1em plus 0.5em minus 0.4em\relax IEEE, 2025, pp. 1276--1281.

\bibitem{WGAN}
M.~Arjovsky, S.~Chintala, and L.~Bottou, ``Wasserstein gan,'' \emph{arXiv preprint arXiv:1701.07875}, 2017.

\bibitem{hua2019gan}
Y.~Hua, R.~Li, Z.~Zhao, X.~Chen, and H.~Zhang, ``Gan-powered deep distributional reinforcement learning for resource management in network slicing,'' \emph{IEEE Journal on Selected Areas in Communications}, vol.~38, no.~2, pp. 334--349, 2019.

\bibitem{gan2018}
A.~Creswell, T.~White, V.~Dumoulin, K.~Arulkumaran, B.~Sengupta, and A.~A. Bharath, ``Generative adversarial networks: An overview,'' \emph{IEEE signal processing magazine}, vol.~35, no.~1, pp. 53--65, 2018.

\bibitem{malektaji2021deep}
S.~Malektaji, A.~Ebrahimzadeh, H.~Elbiaze, R.~H. Glitho, and S.~Kianpisheh, ``Deep reinforcement learning-based content migration for edge content delivery networks with vehicular nodes,'' \emph{IEEE Transactions on Network and Service Management}, vol.~18, no.~3, pp. 3415--3431, 2021.

\bibitem{zhao2020novel}
L.~Zhao, Y.~Liu, A.~Y. Al-Dubai, A.~Y. Zomaya, G.~Min, and A.~Hawbani, ``A novel generation-adversarial-network-based vehicle trajectory prediction method for intelligent vehicular networks,'' \emph{IEEE Internet of Things Journal}, vol.~8, no.~3, pp. 2066--2077, 2020.

\bibitem{seo2018gids}
E.~Seo, H.~M. Song, and H.~K. Kim, ``Gids: Gan based intrusion detection system for in-vehicle network,'' in \emph{2018 16th Annual Conference on Privacy, Security and Trust (PST)}.\hskip 1em plus 0.5em minus 0.4em\relax IEEE, 2018, pp. 1--6.

\bibitem{ioannou2026gan}
I.~Ioannou and V.~Vassiliou, ``Generative adversarial networks for energy-aware iot intrusion detection: Comprehensive benchmark analysis of gan architectures with accuracy-per-joule evaluation,'' \emph{Sensors}, vol.~26, no.~3, p. 757, 2026.

\bibitem{liao2025diffusiondrive}
B.~Liao, S.~Chen, H.~Yin, B.~Jiang, C.~Wang, S.~Yan, X.~Zhang, X.~Li, Y.~Zhang, Q.~Zhang \emph{et~al.}, ``Diffusiondrive: Truncated diffusion model for end-to-end autonomous driving,'' in \emph{Proceedings of the Computer Vision and Pattern Recognition Conference}, 2025, pp. 12\,037--12\,047.

\bibitem{de2022vehicles}
M.~{\'A}. De~Miguel, J.~M. Armingol, and F.~Garcia, ``Vehicles trajectory prediction using recurrent vae network,'' \emph{IEEE Access}, vol.~10, pp. 32\,742--32\,749, 2022.

\bibitem{lyu2026enabling}
N.~Lyu, J.~Zhang, Z.~Shen, C.~Chen, Z.~Jiang, Q.~Chen, and M.~Guo, ``Enabling client-autonomous training optimizations for efficient federated learning,'' \emph{IEEE Transactions on Parallel and Distributed Systems}, 2026.

\bibitem{cha2024multi}
N.~Cha and L.~Chang, ``Multi-objective distributed client selection in federated learning-assisted internet of vehicles,'' \emph{Sensors}, vol.~24, no.~13, p. 4180, 2024.

\bibitem{lu2025clustered}
S.~Lu, H.~Yin, P.~Wang, T.~Zhu, C.~Dai, S.~Dai, and B.~Guo, ``Clustered federated learning with adaptive pruning for 6g edge-intelligent transportation,'' \emph{IEEE Transactions on Intelligent Transportation Systems}, 2025.

\bibitem{nie2025mobility}
X.~Nie, C.~Wang, T.~Zhou, Q.~Zhou, X.~Zhu, and J.~Zhang, ``Mobility-aware cooperative caching in iovs based on secure asynchronous federated and deep reinforcement learning,'' \emph{IEEE Internet of Things Journal}, 2025.

\bibitem{rappaport2002wireless}
T.~S. Rappaport, ``Wireless communications--principles and practice, (the book end).'' \emph{Microwave Journal}, vol.~45, no.~12, pp. 128--129, 2002.

\bibitem{fenn2010performance}
M.~Fenn, J.~Holmes, and J.~Nucciarone, ``A performance and cost analysis of the amazon elastic compute cloud (ec2) cluster compute instance,'' \emph{White Paper}, vol.~10, 2010.

\end{thebibliography}
\end{document}